\documentclass{article}
\usepackage{amssymb}

\usepackage[bookmarks=true]{hyperref}
\usepackage{amsmath,amssymb,amsfonts}
\usepackage{float}
\usepackage{graphicx,subcaption}
\usepackage{stfloats}
\usepackage{booktabs}
\usepackage{multirow}
\usepackage{bm}
\usepackage{pifont}
\usepackage{tabularx}
\usepackage{adjustbox}
\usepackage{color}
\usepackage{enumitem}
\usepackage{color}
\usepackage{tcolorbox}

\usepackage[commandnameprefix=always]{changes}
\setdeletedmarkup{\textcolor{red}{\sout{#1}}}
\definechangesauthor[color=blue]{AM}
\definechangesauthor[color=red]{FG}
\definechangesauthor[color=brown]{VS}

\usepackage[preprint]{corl_2025} 

\title{Zero-shot Interactive Perception}

\author{
  Venkatesh Sripada \\
  University of Surrey \\ 
  \texttt{v.sripada@surrey.ac.uk} 
  \AND
  Frank Guerin \\
  University of Surrey \\
  \texttt{f.guerin@surrey.ac.uk} 
  \And
  Amir Ghalamzan \\
  University of Sheffield \\
  \texttt{a.ghalamzan@sheffield.ac.uk} 
}

\begin{document}
\maketitle

\begin{abstract}
    Interactive perception (IP) enables robots to extract hidden information in their workspace and execute manipulation plans by physically interacting with objects and altering the state of the environment—crucial for resolving occlusions and ambiguity in complex, partially observable scenarios. We present \textit{Zero-Shot IP (ZS-IP)}, a novel framework that couples multi-strategy manipulation (pushing and grasping) with a memory-driven Vision Language Model (VLM) to guide robotic interactions and resolve semantic queries. ZS-IP integrates three key components: (1) an Enhanced Observation (EO) module that augments the VLM’s visual perception with both conventional keypoints and our proposed \textit{pushlines} —a novel 2D visual augmentation tailored to pushing actions, (2) a memory-guided action module that reinforces semantic reasoning through context lookup, and (3) a robotic controller that executes pushing, pulling, or grasping based on VLM output. Unlike grid-based augmentations optimized for pick-and-place, pushlines capture affordances for contact-rich actions, substantially improving pushing performance. We evaluate ZS-IP on a 7-DOF Franka Panda arm across diverse scenes with varying occlusions and task complexities. Our experiments demonstrate that ZS-IP outperforms passive and viewpoint-based perception techniques such as Mark-Based Visual Prompting (MOKA), particularly in pushing tasks, while preserving the integrity of non-target elements.
\end{abstract}



\section{Introduction}
	

Robots are increasingly being deployed in semi-structured environments, dealing with a wide array of unplanned objects and configurations. For example, picking varied items in an e-commerce warehouse, packing boxes, and processing customer returns may involve searching for specific information. 
A critical challenge in robotic perception arises when robots operate in cluttered or partially observable environments, where objects may be occluded or entirely hidden from view \citep{dellaert2017factor}. Traditional approaches, such as active perception \citep{bajcsy1988active}, address this issue by dynamically adjusting the robot's viewpoint to gather additional information. However, sometimes it is necessary to move obstacles out of the way, and this is called interactive perception \citep{bohg2017interactive,gupta2012interactive}


Recent studies have integrated the manipulation capability with high-level reasoning~\citep{fangandliu2024moka} but are limited to performing only pick-and-place or grasping operations. 
Also, they remain constrained to static-scene tasks (e.g., “Insert the perfume into the gift box” or “Unplug the cable”)~\citep{fangandliu2024moka}. Such approaches cannot accommodate queries that depend on temporal context. For instance, “What is under the eraser?” requires the system to track the eraser’s original location before and after manipulation to terminate the search.
Conventional methods are also limited in terms of generalisation to unseen objects~\citep{maactive,kirillov2023segment}.  The ability to interpret natural language queries, reason about object relationships, and plan physical interactions in a cohesive framework 
is still underdeveloped.

We present ZS-IP, a Zero-Shot Interactive Perception framework that addresses these shortcomings by tightly integrating vision-language models, enhanced visual affordances, and memory-driven action planning. Our key contributions are:
\begin{itemize}[leftmargin=*]
    \item \textbf{Zero-shot Interactive Perception:} A novel architecture using a vision-language model to interpret natural-language queries, reason about object relationships, and guide multi-strategy manipulations
    \item \textbf{Pushlines $\&$ Enhanced Observation}: Introduction of pushlines, a 2D visual augmentation that identifies feasible push or pull trajectories, alongside a multi-resolution 2D grid and grasp keypoints (Fig \ref{fig:combined_figures}), enabling a rich set of physical interactions and dynamic camera re-positioning.
    \item \textbf{Memory-Guided Action Module}: A retrieval-based mechanism that summarizes temporal information across sequential states, ensuring precise, context-aware action selection and termination.
    \item \textbf{Comprehensive Evaluation}: Deployment of ZS-IP on a 7-DOF Franka Panda arm across eight interactive perception tasks with varied occlusion and complexity. We demonstrate that ZS-IP significantly outperforms passive and viewpoint-based baselines.
\end{itemize}



\section{Related works}

Recent advances in large language models (LLMs)~\citep{anil2023palm} and vision-language models (VLMs)~\citep{shen2021much} have significantly improved high-level reasoning in robotics. However, spatial reasoning remains a core limitation for VLMs ~\citep{wu2023visual, chen2024spatialvlm}, largely due to insufficient training data involving object relationships and interactions. Efforts like NEWTON~\citep{wang2023newton}, PROST~\citep{aroca2021prost}, and PIQA~\citep{bisk2020piqa} address this gap by exploring physical attributes and affordance reasoning, while SayCan~\citep{ahn2022can}, Code as Policies~\citep{liang2023code}, and Tidybot~\citep{wu2023tidybot} integrate LLMs with robotic control through action grounding and in-context learning.

Despite these strides, many of these methods struggle in dynamic or partially observable environments where physical interaction is essential. Active perception techniques~\citep{bajcsy1988active, kwon2024groundedcommonsensereasoning} enhance scene understanding by changing viewpoints, but rely on static, pre-collected observations, limiting adaptability. Though models like Neural Graspness Fields~\citep{maactive} reduce uncertainty through next-best-view planning, they do not change the  environment to access hidden information.

Interactive perception addresses this limitation by enabling robots to alter the environment directly through actions like pushing or lifting~\citep{gupta2012interactive}. Prior works such as MOKA~\citep{fangandliu2024moka} and PIVOT~\citep{nasiriany2024pivot} have leveraged visual prompting and 2D spatial annotations to support such tasks. However, these approaches rely on discrete annotations and struggle in cluttered scenes. In contrast, our proposed method uses a continuous 2D grid for more expressive physical reasoning, enhancing success rates in complex, real-world queries.

While recent frameworks~\citep{hornung2014mobile,novkovic2020object} have introduced dynamic interaction strategies, they often lack semantic integration, limiting their applicability to predefined object categories. Our work bridges this gap by combining interactive perception with high-level semantic reasoning for generalizable, goal-directed robotic manipulation in unstructured environments.

\begin{figure*}[tb]
    \centering
    \includegraphics[width=\textwidth]{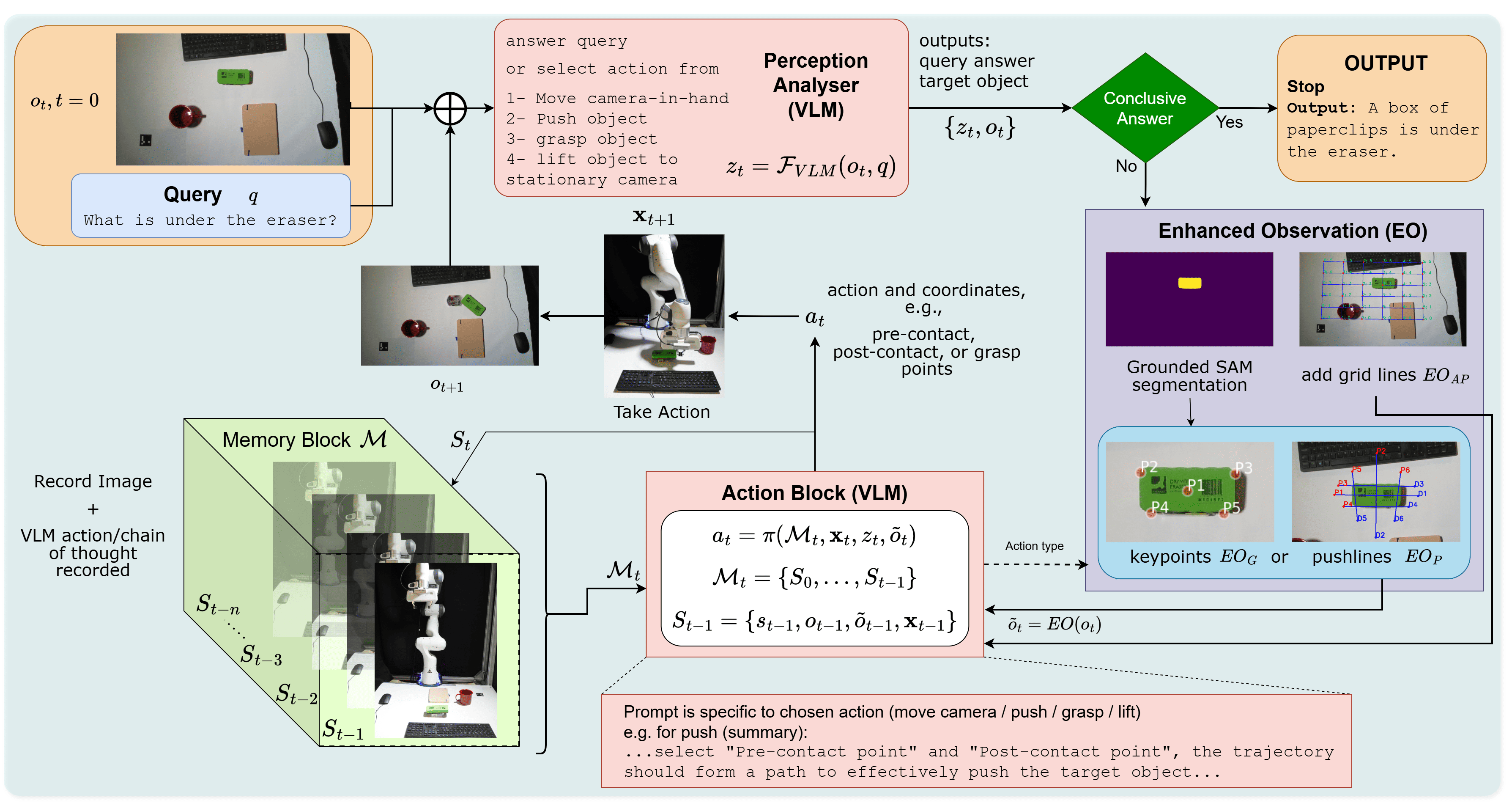}
    \caption{\textbf{Zero-shot Interactive Perception Framework:} The Perception Analyser (i.e. $z_t = \mathcal{F}_{VLM}({o}_t, q)$) evaluates the scene observation $o_t$ at robot configuration $\textbf{x}_t$ and attempts to resolve the query, $q$. If unsuccessful, EO annotates the image, $\tilde{o}_t = EO(o_t)$, after segmenting target objects with Grounded SAM, e.g. push lines ($EO_{\text{P}}$), keypoints ($EO_{\text{G}}$), and {2D} grid ($EO_{\text{AP}}$). A history of interactions is kept in memory $\mathcal{M}_t$ to avoid redundant actions and enhance task efficiency. Action $a_t = \pi( \mathcal{M}_t, \textbf{x}_t, z_t, \tilde{o}_t)$ is generated to make a proper physical interaction to answer the query. $\mathcal{M}_t$ states, $S_t$ contains both images $o_t$ and a summary of scene description, $s_t$, relevant to query.  The robot performs the interaction and saves the corresponding image and robot's state in $S_{t+1}$ which then goes to the Perception Analyser. }
    \label{fig:method_RL}
\end{figure*}

The challenge of integrating semantic reasoning with physical interaction remains a critical bottleneck in robotic perception, affecting applications ranging from inventory assessment of returned packages to delicate object handling, such as clustered strawberry picking. While LLMs and VLMs excel at high-level semantic reasoning, their ability to guide low-level physical interactions remains limited. Efforts like SpatialVLM~\citep{chen2024spatialvlm} have sought to enhance spatial representations, but they primarily rely on static annotations and lack the capability to interact dynamically with their environment. Similarly, physics-based commonsense reasoning frameworks \citep{aroca2021prost,wang2023newton} focus on object properties and interactions but do not fully leverage interactive perception to iteratively manipulate and reveal occluded information.
%


\section{Methodology}

Our system, comprising a robotic manipulator equipped with an in-hand camera (Fig.~\ref{fig:combined_figures}), acts as an \textit{agent} that interacts with its environment to reveal hidden or occluded objects and answer specific queries. The core task involves analysing the scene, performing physical interactions to change the workspace's state, and iteratively gathering visual data until the VLM can provide a conclusive answer to a query, e.g. `\emph{What is under the eraser?}' in Fig.~\ref{fig:combined_figures} bottom row. 

Given a natural language query, $q$, and an initial visual observation $o$ at time $t$, the agent uses a \textit{Perception Analyser} to output an answer $z$, $z_t = \mathcal{F}_{VLM}({o}_t, q)$ (see schematic of ZS-IP in Fig.~\ref{fig:method_RL}). If the answer is not conclusive, ZS-IP generates an action $a_t = \pi( \mathcal{M}_t, \textbf{x}_t, z_t, \tilde{o}_t)$ to change the scene configuration aiming to maximise the likelihood of answering the query in the next iteration. This is detailed below:


\noindent \textbf{Input Block:} It contains the initial visual data from the camera-in-hand image $o_t, t=0$ (see Fig.~\ref{fig:combined_figures}), taken at the robot's home configuration, $\textbf{x}_t$, and the natural language query (e.g., `What is under the eraser?') $q$. This provides the system with the starting conditions for exploration.

\subsection{Perception Analyser (PA)} It interprets visual data and assesses whether the current observation is sufficient to answer the query, $z_t = \mathcal{F}_{VLM}({o}_t, q)$. $z_t$ includes the evaluation of the scene for objects relevant to $q$, spatial relationships, and context, and a binary decision: either the query is resolved, or further workspace interaction is required.  

\subsection{Enhanced Observation:}  
We introduce the Enhanced Observation (EO), $\tilde{o}$, which may include the image with overlaid 2-D grid, pushing lines or grasping key points, see Fig.~\ref{fig:combined_figures}, top. A segmentation mask, generated using Grounded SAM~\citep{ren2024grounded}, is used to segment the target object based on a text string descriptor passed by the Perception Analyser (VLM). The mask is provided as input to generate push lines or keypoints
This enhances the VLM to convey its predictive outputs to the robot in the form of planned motions.

\noindent \emph{{Push lines $EO_P$:}} The agent uses Principal Component Analysis (PCA) \citep{pca1901} and computes Principal axis 1 and 2 based on the object segmentation mask
Two principal push lines are generated along the axes passing through the centre of the object segmentation mask and oriented perpendicular to the object's shape (Fig.~\ref{fig:combined_figures} top right). Additionally, four more push lines are created using key points near the edges of the object's segmentation mask. These push lines are annotated as \textit{P} (pre-contact points) and \textit{D} (post-contact points), followed by indices (e.g., \textit{P1-P6} and \textit{D1-D6}). These points define the start and end locations of a virtual push trajectory.

\noindent \emph{{Grasping key points  $EO_G$:}} We define key points as a set of five points, with four points marking the object boundary (from the segmentation mask) and the fifth being the centroid. The key points are generated using Farthest Point Sampling (fps) \citep{qi2017pointnet}. The key points are overlaid on the image of the segmented target object (Fig.~\ref{fig:combined_figures} bottom raw) with indices \textit{P}, i.e.~\textit{P1-P5}.

\noindent \emph{Virtual Grid $EO_{ap}$}: a 2-D virtual grid is built in the robot workspace and projected into the image frame using an ArUco marker $M$ anchoring the grid. Its vertices $V^{M}_{v}$ are defined in the camera coordinate frame $C$. 
The transformation of these vertices into the robot's base frame $B$ is computed using a homogeneous transformation matrix $T$:

\[
V_{v}^{B} = T_{C}^{B} T_{M_{id}}^{C} V_{v}^{M_{id}}
\]

Here, $T_{M_{id}}^{C}$ is derived from the camera's intrinsic parameters using translation and rotation vectors, while $T_{C}^{B}$ is calculated through the robot’s inverse kinematics. 
The system continuously updates the translation and rotation vectors as the robot moves, maintaining a consistent grid projection. The agent internally generates a fine grid and zooms where precise coordinates for the robot's movements are required.
The distance from the camera to the objects is calculated using data from the depth camera. The geometric centre of the segmentation mask is identified, and the depth information corresponding to this point is retrieved and used where needed.

\subsection{Memory-guided action module}

It consists of robot state representations and memory block that are provided as input to a VLM. The VLM generates an action policy.

\noindent \textbf{State Representation:}  
The experiment's state at time $t$ captures current configuration of the robot, $\textbf{x}_t$, $o_t$, $\tilde{o}_t$ and the workspaces' state, $s_t$, i.e. summary information of the scene after each interaction, as well as the VLM chain of thought at that time, the states visited, robot trajectory at time $t$ and the context and knowledge, such as task requirements, constraints and goals.

\noindent \textbf{Memory Block:} 
%
Information gathered during each movement is stored and used iteratively to refine subsequent actions, ensuring that the robot accumulates relevant knowledge for query resolution. This consists of memory data $\mathcal{M}_t = \{S_0, \dots, S_{t-1}\}$, $S_t = \{s_t, o_t,\tilde{o}_t, \textbf{x}_{t-1}\}$. This information is retrieved and comprehensively provided as context enabling the robot to dynamically and intelligently interact with its environment, uncover hidden objects, and resolve queries effectively. 

The ZS-IP Policy $(\pi)$ first decides the proper type of interaction.
It then utilises $o(s_t)$, $\tilde{o}_t$ and memory $\mathcal{M}_t$ to select a precise trajectory of motion, $a_t = \pi( \mathcal{M}_t, \textbf{x}_t, z_t, \tilde{o}_t)$. The policy balances exploration and task objectives by guiding the robot to actions that maximise the likelihood of answering the query. We transform the actions from the camera frame $C$ to the robot's base frame $B$. The robot transitions between configurations $q_t$ and $q_{t+1}$ for movements like reach-to-grasp/push (using OMPL planner) whereas in pushing movements the robot follows a continuous trajectory (using Pilz Industrial Motion Planner).

\subsection{Robot interaction} We categorised robot actions to evaluate the contribution of each mode of interaction ($a_1$)~move camera-in-hand, ($a_2$)~push, ($a_3$)~grasp (pick-and-place) and ($a_4$)~lift to a stationary camera to investigate. 

\begin{figure}[t!]
    \centering

    \begin{minipage}[t]{0.51\textwidth}
        \centering
        \includegraphics[width=0.48\linewidth]{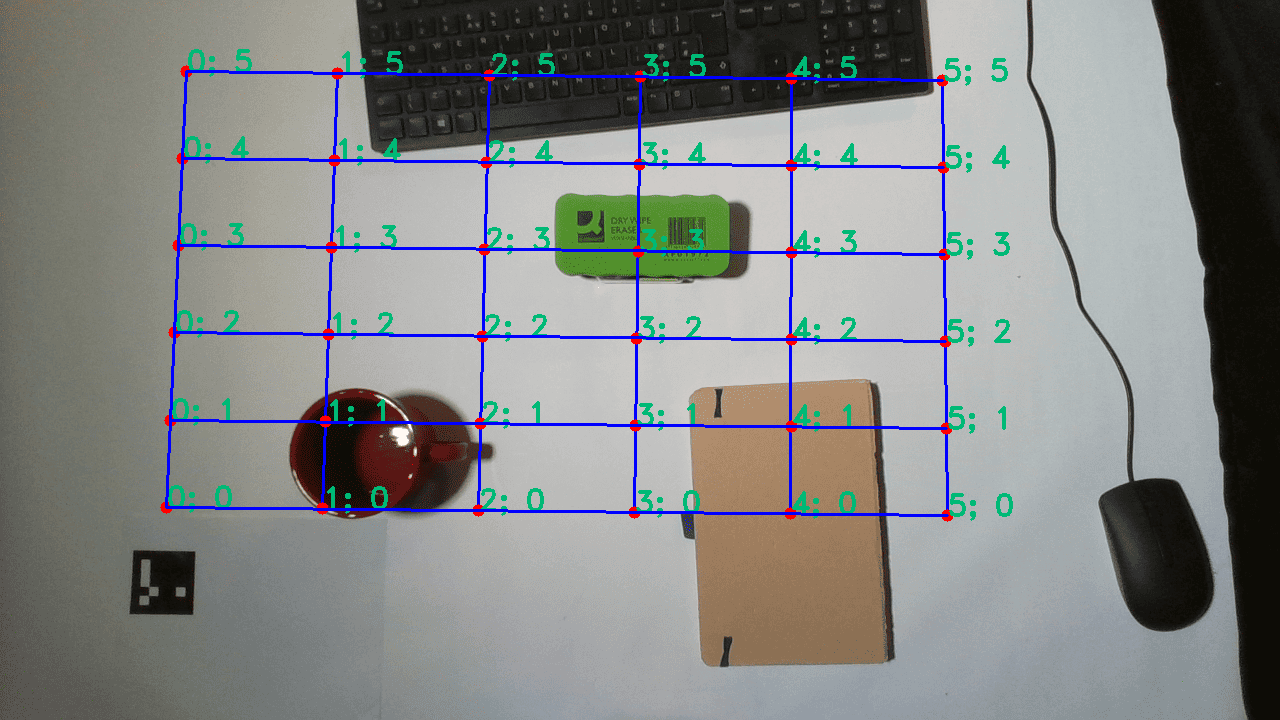}
        \includegraphics[width=0.48\linewidth]{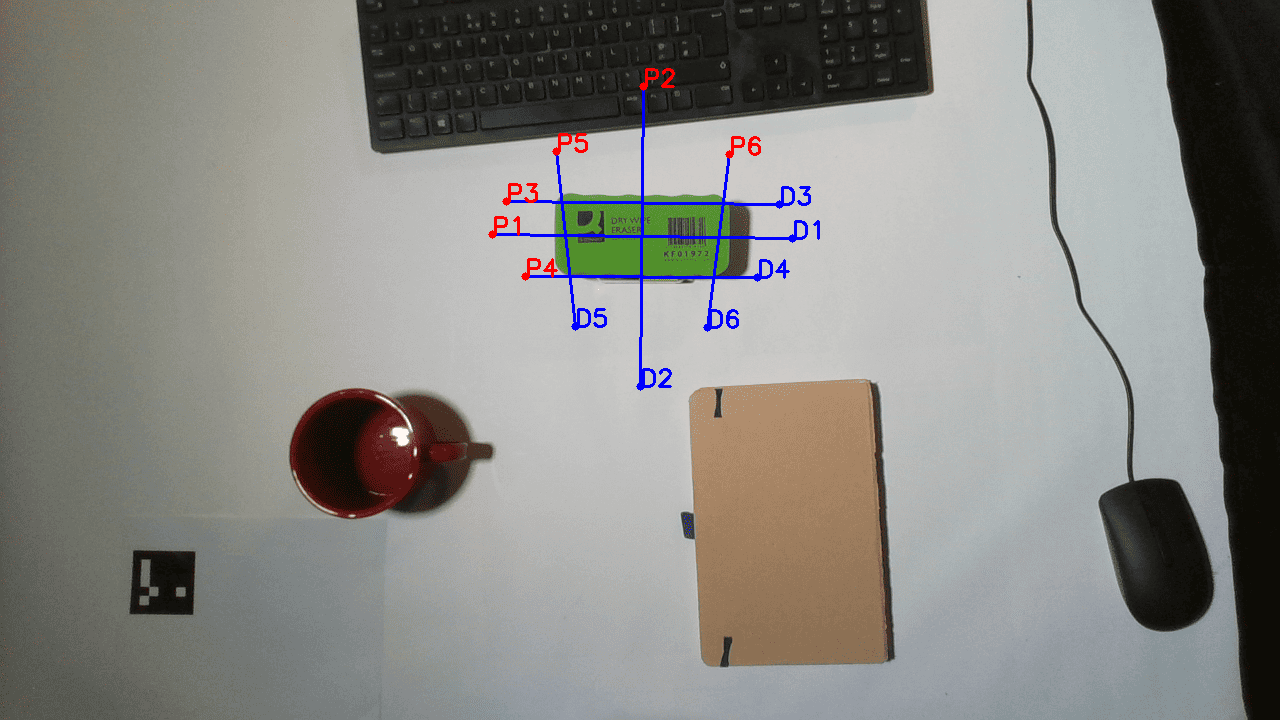}
        \\
        \includegraphics[width=0.48\linewidth]{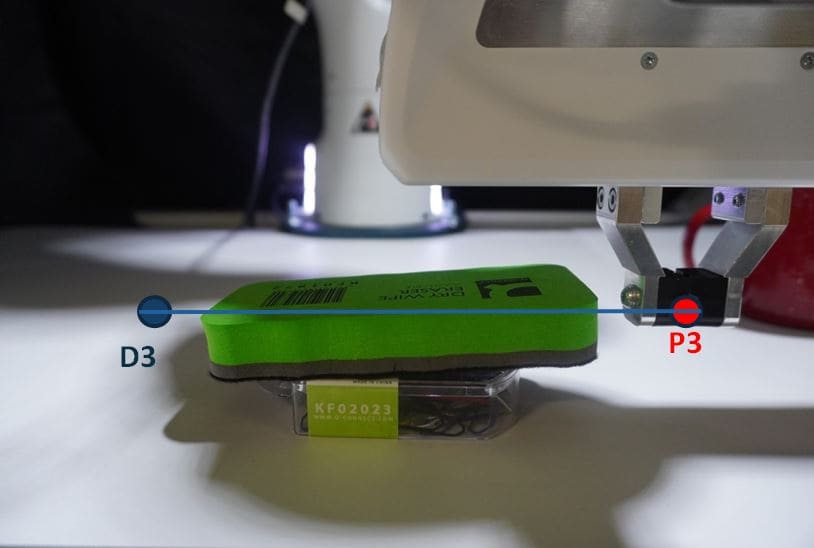}
        \includegraphics[width=0.48\linewidth]{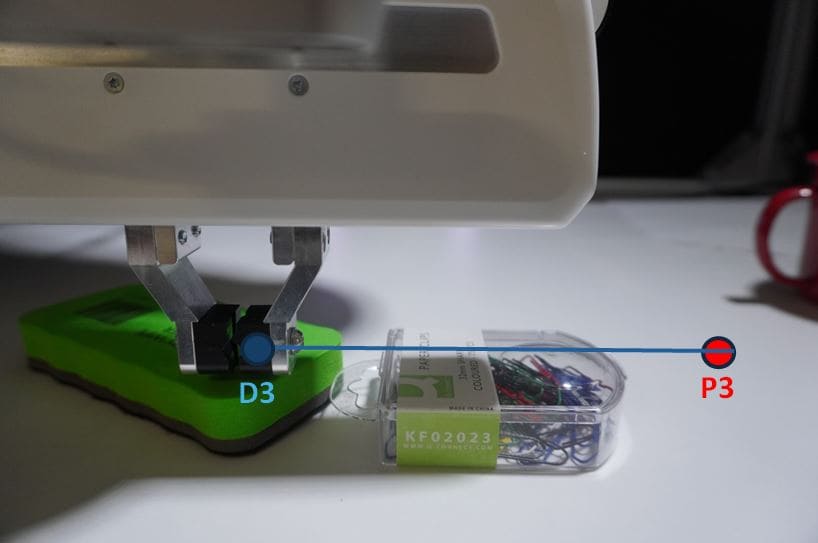}
        \caption*{Push lines}
    \end{minipage}
    \hfill
    \begin{minipage}[t]{0.45\textwidth}
        \centering
        \includegraphics[width=0.48\linewidth]{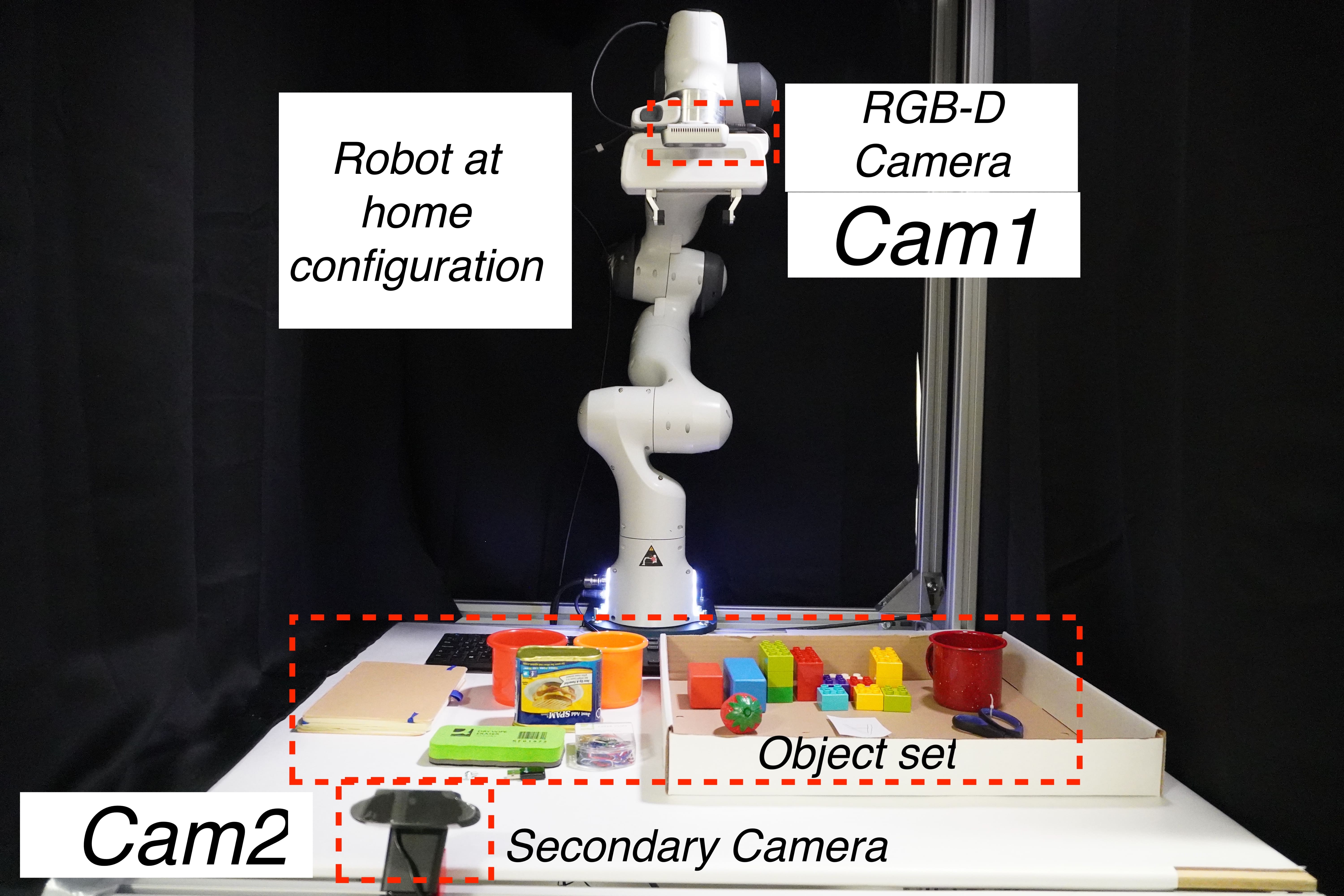}
        \includegraphics[width=0.48\linewidth]{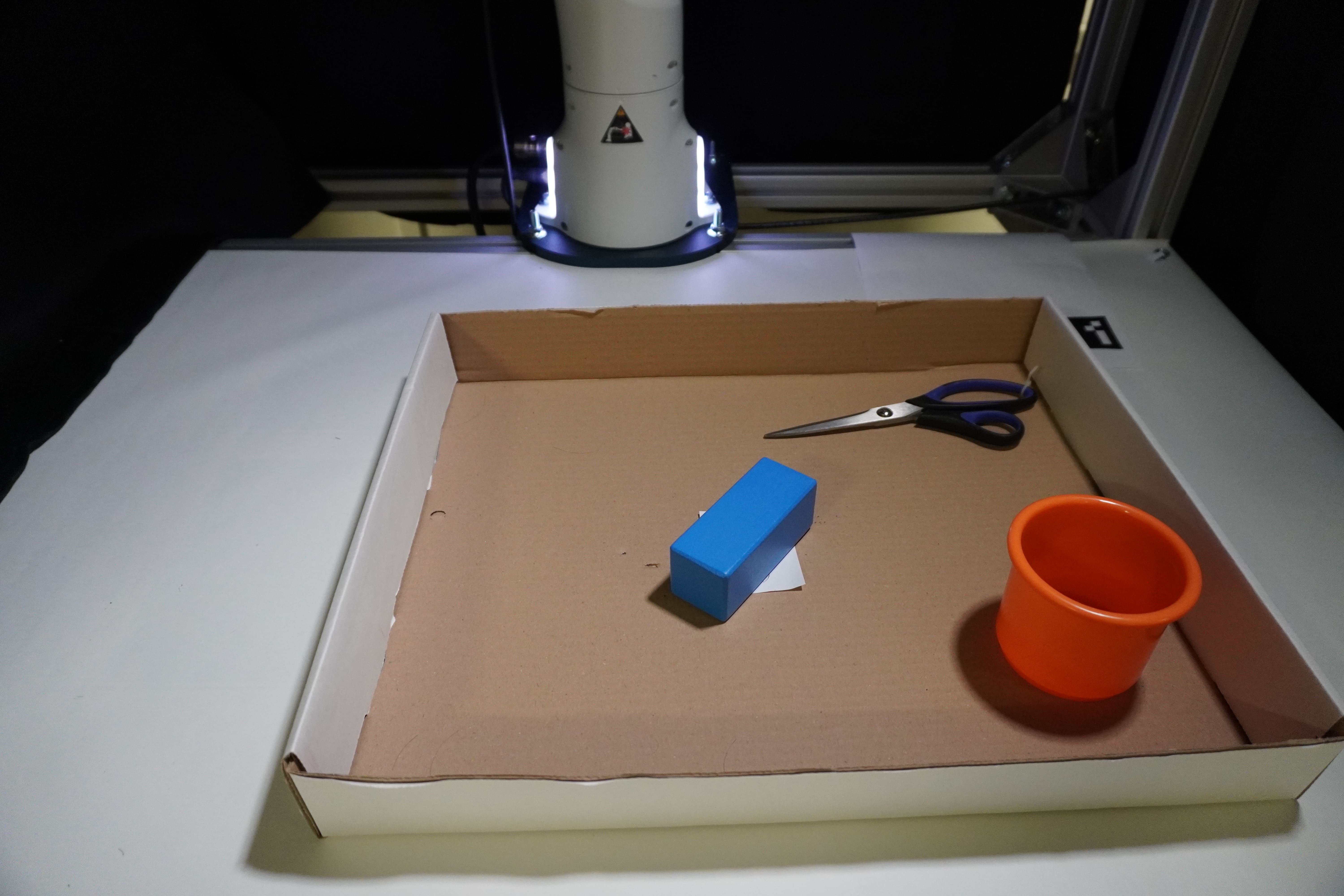}
        \\
        \includegraphics[width=0.48\linewidth]{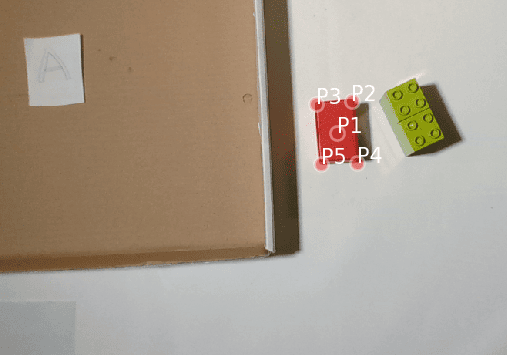}
        \includegraphics[width=0.48\linewidth]{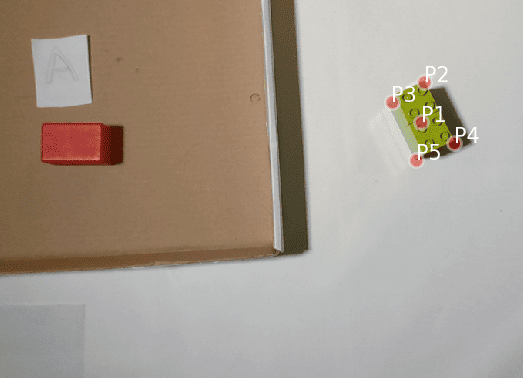}
        \caption*{Grasp keypoints}
    \end{minipage}
    
    \caption{\textit{Four 2x2 images to the left show Push lines:}(Top left and right) A virtual 2D grid overlaid on the table, projected onto the 2D image, along with push lines generated for Task I (Query: What is under the eraser?). The robot selects P3 as the pre-contact point (bottom left) and D3 as the post-contact point (bottom right) to execute the pushing action. This interaction successfully uncovers and reveals the box of paper clips previously hidden underneath.
    \textit{Four 2x2 images to the right show setup and Grasp keypoints:} Top: Robotic setup (left), scene for task VI; Bottom: key points generated for two objects in Task V involving pick-and-place at time step 1 (left) and 2 (right).
    }
    \label{fig:combined_figures}
    \vspace{-.2cm}
\end{figure}

\noindent ({${a}_1$}) the robot simply moves the camera on its wrist to improve its view of the target object. The robot adjusts the orientation of its end-effector based on the object's shape and orientation, enabling it to look into or beside the object to gain better visibility. 
({${a}_2$}) the robot performs pushing actions. $\pi$ analyses the push line trajectories (added by enhanced observation) and selects the optimal push line for the task. 
({${a}_3$}) Grasping actions are guided by key points generated by enhanced observation. $\pi$ selects the most feasible key point as the \textit{grasp location}, indicating where the object should be grasped. Additionally, $\pi$ determines a \textit{place location}, representing the target position for placing the object after grasping.  
($a_4$) The robot first executes the grasping action using the grasping key points, \emph{$EO_G$}. It then positions the object in front of the secondary camera, which is fixed on the table (Fig.~\ref{fig:combined_figures}, top left). The robot holds the object until an image is captured, ensuring that it can accurately answer the query $q$, e.g., \textit{“What is written under the eraser?”}

All the key points, push lines and 2-D grid, are mapped into the robot's base frame. For improved precision, we use zoomed-in images for ZS-IP to select the key points or push lines.

    
    
    

\noindent \textbf{Iterative Process:}  
The robot starts in a home configuration, capturing an initial observation $o_0$. The \textit{Perception Analyser} evaluates observation and generates a response $z_t = \mathcal{F}_{VLM}({o}_t, q)$. If a conclusive answer is not generated, \textit{Enhanced Observation} generates the corresponding annotated images and passes it to Action Block that generates action $a_t = \pi( \mathcal{M}_t, \textbf{x}_t, z_t, \tilde{o}_t)$ using the \textit{Memory} state, $\mathcal{M}_t$, to generate an action $a_t$, which changes the workspace and acquires a new observation $o_{t+1}$. The updated observation $o_{t+1}$ is fed back into the Perception Analyser, $z_{t+1} = \mathcal{F}_{VLM}({o}_{t+1}, q)$ continuing the iterative process until the query is resolved or the maximum iteration limit is reached.

By combining interactive perception with semantic reasoning through VLMs, the proposed framework allows robots to manipulate their environments iteratively, revealing occluded information and improving task performance 


\section{Experiment Setup and Structure}
\label{sec:exp_setup}

\noindent \emph{Robotic Setup:}  
The experiments were conducted using a 7-DOF Franka Emika Panda arm. The robot was equipped with an Intel RealSense RGB-D camera (Cam1) mounted on its end-effectors and another RGB camera (Cam2) mounted on the frame attached to the table (see Fig.~\ref{fig:combined_figures} top left). The robot starts from the home configuration, and the RGB input taken from Cam1 is utilised to analyse spatial relationships, while depth information was leveraged to estimate the object's distance. 

To enhance spatial localization and precise action execution, we placed an \emph{ArUco marker} within the robot’s workspace. The marker served as an anchor for generating the 2D virtual grid. In future, we plan to remove the marker from our framework. The workspace included objects selected from the \emph{YCB} object set~\citep{calli2015ycb}, ensuring diverse testing scenarios with varying levels of complexity and occlusion. 
We integrated the GPT-4o model as the Perception Analyser to interpret visual inputs and guide the robot’s actions. 
The VLM temperature parameter was set to zero to increase the consistency and reproducibility of results.


\noindent \emph{Experimental Design:}  
To evaluate the ZS-IP framework, we designed experiments with varying task complexities and scenarios. The following aspects were analysed: (\emph{i}) Comparing interactive perception to baseline scenarios with static observations (i.e., passive perception).  
(\emph{ii}) Evaluating different modes of workspace interaction, including camera-in-hand movement, pushing objects, grasping objects, and lifting objects for further investigation.  
(\emph{iii}) Assessing the influence of enhanced observation, e.g., 2D grid-based augmented annotation, augmented key points, and pushing lines, on decision-making and performance.  
(\emph{iv}) Measuring the framework’s adaptability to complex environments with partially or fully occluded objects.

We designed eight tasks with varying levels of complexity to test all four action categories. 
The ZS-IP chooses suitable action categories across all tasks based on the contextual requirements of the scene. Each task was executed 10 times with slight variations in positioning and orientation of objects.  Results are reported in Table~\ref{tab:ip_vlm_sr}. The prompts and corresponding images are listed out in Appendix \ref{sec:tasks_prompts} while Fig.~\ref{fig:enter-label} shows tasks I (top), II (middle) and III (bottom row).


\noindent \emph{Performance Metrics}:
{We adopted and modified evaluation metrics inspired by the R2R-VLN dataset~\citep{anderson2018vision}, tailoring them for our interactive manipulation tasks:} \emph{Success Rate} (\textbf{SR}): The percentage of trials in which the robot could correctly give a \emph{conclusive answer to the query}. 
{\emph{Total Length (Position)}} (\textbf{TL}): The mean total distance travelled by the robot across all trials. {\emph{Total Length (Position, Successful)}} (\textbf{TLS}): The mean total distance travelled by the robot across successful trials. 
{\emph{Position Error}} (\textbf{PE}): The mean Euclidean distance between the robot's final position and the target position.
{\emph{Oracle Success Rate}} (\textbf{OSR}): The success rate when the robot stops at the closest point (within a margin of $0.1\: m$) to the goal point annotated by a human.

We also compared ZS-IP with MOKA, which is a state-of-the-art motion generation framework for open-world robotic manipulation. MOKA uses point-based affordance representation to bridge the gap between VLM's prediction on RGB images and robot's motion ~\citep{fangandliu2024moka}. For a fair comparison we adapt the output with Enhanced Observation ($EO$) while retaining the functionality of the other components.

This comprehensive experimental setup allows us to evaluate the effectiveness of the proposed Interactive Perception framework in dynamically modifying the workspace and resolving queries involving partially or fully hidden objects.


\section{Results and Discussion}
\emph{Task Success Rates:} ZS-IP achieved high success rates on less complex tasks (Tasks I, II, and IV), demonstrating strong reasoning, perception, and memory-guided manipulation capabilities. In Task I, ZS-IP successfully planned and executed object interactions; in Task II, it autonomously altered lighting conditions by moving a tin to reveal a hidden object. In Task IV, it leveraged temporal memory to track a block's position and expose hidden text beneath it.

Task III was more challenging, requiring precise pushing to retrieve a pen drive buried under clutter. Failures stemmed from delicate manipulation demands and exceeding the five-iteration limit. Similarly, Task V involved placing two blocks in a box while preserving their spatial relationship—ZS-IP performed well (SR = 0.8), with failures due to errors in spatial reasoning.

Task VI required lifting an eraser to read its surface text. ZS-IP used a secondary camera effectively, achieving a success rate of 0.8. Across tasks, the system showed resilience to positional and orientational variance, validating its generalisation to unstructured scenes.

Notably, performance in more complex tasks (e.g., Task III and Task VII) highlighted the framework's reasoning, memory recall, and precise sequencing of combined actions. These results showcase its robustness and potential for adaptation to more complex scenes in unstructured environments.
We also observed that ZS-IP demonstrated robustness to variations in object position and orientation, successfully completing tasks despite changes in spatial configuration.

\begin{table*}[t!]
\centering
\caption{ZS-IP performed for eight different tasks with different required levels of complexity from easy (i.e. Task I) to more complex (Task VIII) interactions. Four categories of interaction are considered: `Pushing', `Pick-and-place' or `Grasping', `Lifting' an object to investigate it, and `observe'. Each task is performed 10 times, and the results show the success rate out of 10 trials. }
\resizebox{\textwidth}{!}{ 
\footnotesize	
\begin{tabular}{lcccccccc}
\toprule
\multirow{4}{*}{\textbf{Method}} & 
\multicolumn{3}{c}{\textbf{Pushing}} & 
\multicolumn{2}{c}{\textbf{Grasping}} & 
\multicolumn{2}{c}{\textbf{Lifting to Investigate}} &
\multicolumn{1}{c}{\textbf{Observe}}\\ 

\cmidrule(lr){2-4} \cmidrule(lr){5-6} \cmidrule(lr){7-8} \cmidrule(lr){8-9} 
 & \textbf{Task I} & \textbf{Task II} & \textbf{Task III} & 
\textbf{Task IV} & \textbf{Task V} & 
\textbf{Task VI} & \textbf{Task VII} & \textbf{Task VIII} \\
\midrule
ZS-IP  & 0.9 & 0.9 & 0.6 & 1.0 & 0.8 & 0.8 & 0.2 & 0.7\\
\bottomrule
\end{tabular}
}
\label{tab:ip_vlm_sr}
\end{table*}

Tasks VII and VIII, representing the highest levels of complexity, were assessed across all five frameworks. Oracle positions for these tasks were determined through human-annotated ground truth labelling, performed post-experimentation for each scene. Each scene incorporated a single predefined push line and a designated key point, both manually specified by an expert annotator. These annotations were carefully selected to minimise scene disruption while ensuring successful task execution.

For the \emph{ZS-IP In-Context}, two annotated images and corresponding instructions were provided before the system was exposed to the query of the current task. This approach aimed to enhance contextual understanding and action selection, improving performance in these more demanding scenarios. To aid semantic understanding of the scene, we introduce a Retrieval Augmented In-context Generation (RAIG) method, explained in the Appendix \ref{appendix:raig}, but is opted out during experimentation.

\emph{Task VII Analysis:} ZS-IP Push failed to resolve \emph{Task VII} (Fig.~\ref{fig:task_7_images} a-c) as the bottom face of the eraser remained occluded, requiring lifting and reorientation for visibility by Cam2. Despite multiple pushing attempts, ZS-IP lacked sufficient task knowledge and returned ‘None’, indicating no viable action. This likely stemmed from VLM hallucination, as it prematurely assumed the eraser text was in English without evidence. While ZS-IP Grasp could theoretically solve the task by relocating surrounding objects, it failed after exceeding the five-iteration limit.

The MOKA framework, despite using both pushing and grasping actions alongside in-context examples, also failed to complete the task. Pushing an object proved to be a challenge for MOKA as the VLM was unable to suggest pre- and post-contact points only based on the discrete 2D grid. It was also limited by the lack of memory retention for prior states and scene configurations. An ablation was performed by comparing PIVOT~\citep{nasiriany2024pivot} (Appendix \ref{appendix:ablation}) on one task each from the `grasping', `lifting', and `Observe' categories and found that ZS-IP has higher SR. We also compare against varying VLMs outside and within OpenAI's GPT and find GPT-4 to have the highest SR.

 \noindent \textbf{Comparison of methods and failure analysis: }
To rigorously evaluate the limits of our framework, we conducted an in-depth analysis of more complex scenarios for Task VII and VIII (presented in Table~\ref{tab:in_depth_results}). Task VII is similar to Task VI but has additional complexity imposed by obstructing objects on both sides of the eraser. This necessitates multi-step reasoning, requiring ZS-IP to push and separate the obstructing objects before grasping eraser. This process involved \emph{sequential decision-making by retrieving past data from memory}.

\begin{table*}[tb!]
\centering
\caption{Comparing ZS-IP with different features and MOKA. The performance Metrics include SR (Success Rate), TL (Total Length) TLP (Total Length Successful), PE (Position Error) and OSR (Oracle Success Rate).}
\label{tab:in_depth_results}
\resizebox{\linewidth}{!}{
\begin{tabular}{|l|c|c|c|c||c|c|c|c|}
\hline
\multirow{2}{*}{\textbf{Method}} & \multicolumn{4}{c||}{\textbf{Task VII}} & \multicolumn{4}{c|}{\textbf{Task VIII}} \\ \cline{2-9}
 & \textbf{SR $\uparrow$} &\textbf{TL, TLS $\downarrow m$} & \textbf{PE} $\downarrow m$ & \textbf{OSR $\uparrow$} & \textbf{SR $\uparrow$} & \textbf{TL,TLS $\downarrow m$} & \textbf{PE $\downarrow m$} & \textbf{OSR} $\uparrow$ \\ \hline
MOKA              & 0.0         & \{3.88, -\}           & 0.40         & 7         & 0.5         & \{1.96, 1.34\}         & 0.60         & 4        \\ \hline
ZS-IP Observe             & -         & -          & -        & -         & 0.1         & \{0.48, 0.35\}         & {0.15}         & 1         \\ \hline
ZS-IP Push            & 0.0         & \{3.25, -\}          & 0.48         & 7         & 0.8         & \{1.48, 1.63\}         & 0.57         & 4         \\ \hline
ZS-IP Grasp            & 0.0         & \{3.04, -\}          & 0.39        & 5         & -         & -        & -         & -         \\ \hline
ZS-IP             & 0.2         & \textbf{\{3.41, 3.08\}}          & \textbf{0.47}        & 7         & 0.7         & \textbf{\{1.47, 1.56\}}         & 0.64         & 4         \\ \hline
\textbf{ZS-IP In-Context}              & \textbf{0.7}         & \{3.48, 3.52\}          & 0.54        & \textbf{8}         & \textbf{0.8}         & \{1.69, 1.41\}         & \textbf{0.57}         & \textbf{5}         \\ \hline

\end{tabular}
}
\end{table*}


\textit{ZS-IP In-Context} demonstrated a significantly higher SR of 0.7, benefiting from its ability to delay the grasping action until the eraser was entirely free from obstructions. Notably, the Total Length (TL) for \textit{ZS-IP In-Context} was higher compared to ZS-IP due to the VLM occasionally recommending the removal of all surrounding obstacles—including the red container and the Lego structure—before attempting a grasp. While this approach deviated from the oracle manipulation strategy, which prioritised direct manipulation of the eraser, it ensured grasping under optimal conditions, thereby reducing execution errors and increasing overall success.

\noindent \emph{Task VIII Analysis:}  
Task VIII (Fig.~\ref{fig:task_7_images}d) involved locating a strawberry inside a cup. However, to introduce additional complexity, a brown book was placed at an incline, partially obstructing the robot's direct line of sight. While all methods, except grasping, managed to handle this scenario, the task required ZS-IP to conduct precise scene analysis to guide the in-hand camera over the target object for visibility. Additionally, within the pushing action category, ZS-IP had to predict and execute a carefully controlled sideways push near the centre to prevent unnecessary disruptions and further complications.


ZS-IP Observe demonstrated notable efficiency in navigating toward the cup, reducing TL, TLS, and PE. However, it successfully identified the strawberry inside the cup only once. Although ZS-IP accurately provided end-effector targets based on $EO_{ap}$ to position the robot near the cup, it failed to correctly classify the red object as a strawberry in six instances. This failure resulted from the shadow cast by the inclined book, which obstructed essential visual cues for reliable object recognition. Even after repositioning the robot closer to the cup, ZS-IP Observe lacked sufficient contextual understanding to effectively disambiguate the occluded scene.

While Task VIII did not heavily rely on memory capabilities, the baseline MOKA method achieved an SR of only 0.5, primarily due to its dependence on a fixed 2D grid. Selecting a grid location near the top or bottom of the book often led to misaligned pre-contact and post-contact points. Consequently, the robot frequently pushed either the cup or the book’s base, causing unintended scene disruptions such as the cup toppling over. These results underscore the limitations of rigid 2D-based approaches in tasks that require precise and adaptive interactions and highlight the advantage of the closed-loop perception-action framework of ZS-IP, as well as the impact of push lines in complex and unstructured environments.


\section{Conclusion}
\vspace{-.1cm}

We introduced ZS-IP, an interactive perception framework that integrates VLMs with robotic systems for dynamic scene exploration. By leveraging semantic segmentation, 2D spatial grids, grasping key points, and push-line annotations, ZS-IP enables robots to iteratively gather visual information and infer the presence of partially or fully occluded objects. 
Our experiments with a 7-DOF robotic arm demonstrated the efficiency of ZS-IP in highly challenging scenarios, where goal objects were entirely hidden and required multi-step reasoning combined with efficient memory retrieval. The comparative analysis with baseline methods, such as MOKA, highlights the superiority of ZS-IP, underscoring the importance of memory retrieval and dynamic observation gathering. 


\section{Limitations}
\vspace{-.1cm}

Despite the advantages of ZS-IP, its performance is constrained by several factors. One notable limitation is the low-resolution depth information, which restricts precise interaction with objects, particularly in cluttered environments where fine-grained manipulation is required. Additionally, integrating large models in real-time for multi-modal reasoning remains an open challenge and a promising avenue for future research.

Future work will focus on extending ZS-IP to more realistic, real-world scenarios beyond controlled laboratory settings. A key application area is agricultural robotics, where tasks such as fruit picking demand the ability to identify and differentiate ripe fruits hidden behind leaves while avoiding unripe ones. Another significant challenge lies in enhancing the common-sense reasoning capabilities of VLMs. For instance, in the task of arranging two wooden blocks next to each other inside a cardboard box, the VLM sometimes incorrectly stacks one block on top of another and classifies the task as \textit{successful}, as observed in the supplementary video. However, from a human perspective, such cases represent failure modes, as documented in Table~\ref{tab:ip_vlm_sr}, Task V. This discrepancy between VLM-driven task evaluation and human common sense highlights the need for improved reasoning mechanisms in VLMs in robotics.

Furthermore, ZS-IP's current manipulation capabilities remain constrained in terms of action representation. At present, grasping is performed using rotation in $SO(2)$ and translation in $\mathbb{R}^3$, while pushing actions are limited to translations in $\mathbb{R}^2$. These constraints restrict the system's ability to manipulate objects in more complex environments where full $SE(3)$ transformations are required. Future work will focus on extending ZS-IP’s pushing and grasping capabilities to enable more adaptive and robust interactions in unstructured settings.

Finally, incorporating additional sensing modalities, such as tactile feedback and proprioceptive sensing, holds great potential for enhancing ZS-IP’s capabilities in physical interaction. These modalities would enable richer environmental feedback, allowing for more precise and dexterous manipulation. By addressing these limitations, we aim to further advance ZS-IP towards real-world deployment in diverse, challenging environments.


\clearpage



\bibliography{references}  

\appendix
\section{Appendix}
\subsection{Retrieval Augmented In-context Generation (RAIG)}
\label{appendix:raig}

Once the output from Perception Analyser is obtained, we move to the augmenting the semantic understanding of the scene using this module.

We utilize \textit{Retrieval-Augmented In-Context Generation} (RAIG) with eight in-context examples sourced from the Visual Reasoning task suite of the VIMA Benchmark dataset~\citep{jiang2023vima}, focusing specifically on enhancing the visual perception capabilities of vision-language models (VLMs).

\noindent \textit{In-context data generation:} 
We re-implement the Visual Reasoning tasks internally using a demonstrator that chooses from five atomic functions: \texttt{reach\_object()}, \texttt{pick\_up()}, \texttt{go\_to()}, \texttt{place()}, and \texttt{push()}. For each execution, we record the end-effector positions and corresponding image frames upon function completion. The collected data is temporally ordered and processed to serve as input to a VLM. The VLM then generates structured outputs, including a query description, a detailed task description, the inferred action type (e.g., \textit{push} or \textit{grasp}), and a rationale for the chosen action. 
All tasks are executed using an impedance-controlled Franka Emika Panda robot, configured with zero-torque and gravity compensation enabled to ensure safe and compliant interaction. This setup is intended to produce high-quality data that supports reproducibility, improves future context generation, and leverages the expected continual improvements in large vision-language models.

\noindent \textit{Context Retrieval:}
Following the methodology outlined in~\citep{lv2024robomp}, we compute embeddings of the VLM-generated outputs from the Perception Analyser and each in-context example. Cosine similarity is used to retrieve the top-$K$ most similar examples. We use SentenceBERT~\citep{reimers2019sentence} to compute embeddings, enabling semantically meaningful comparisons across textual contexts.

\noindent \textit{Action generation:} 
The retrieved $K$ in-context examples are then fed into the VLM. Based on its interpretation of the current image frame and commonsense reasoning, the VLM selects both the target object and the most relevant example. The action type associated with the selected in-context example is then executed. 
This VLM-driven selection dictates both \textit{what} action is performed and \textit{on which} object. Notably, naively executing the top-ranked retrieved example, without further VLM inference, often resulted in incorrect behavior, such as performing a \textit{grasp} instead of a \textit{place}. Moreover, presenting all $K$ examples simultaneously, as done in~\cite{lv2024robomp}, led to a consistent defaulting to the \textit{grasp} action type. We attribute this to the underrepresentation of \textit{pushing} actions in VLM training datasets, which are predominantly composed of pick-and-place interactions.

\subsection{Additional Images}

\begin{figure}[tb!]
    \centering
    \includegraphics[width=0.9\linewidth]{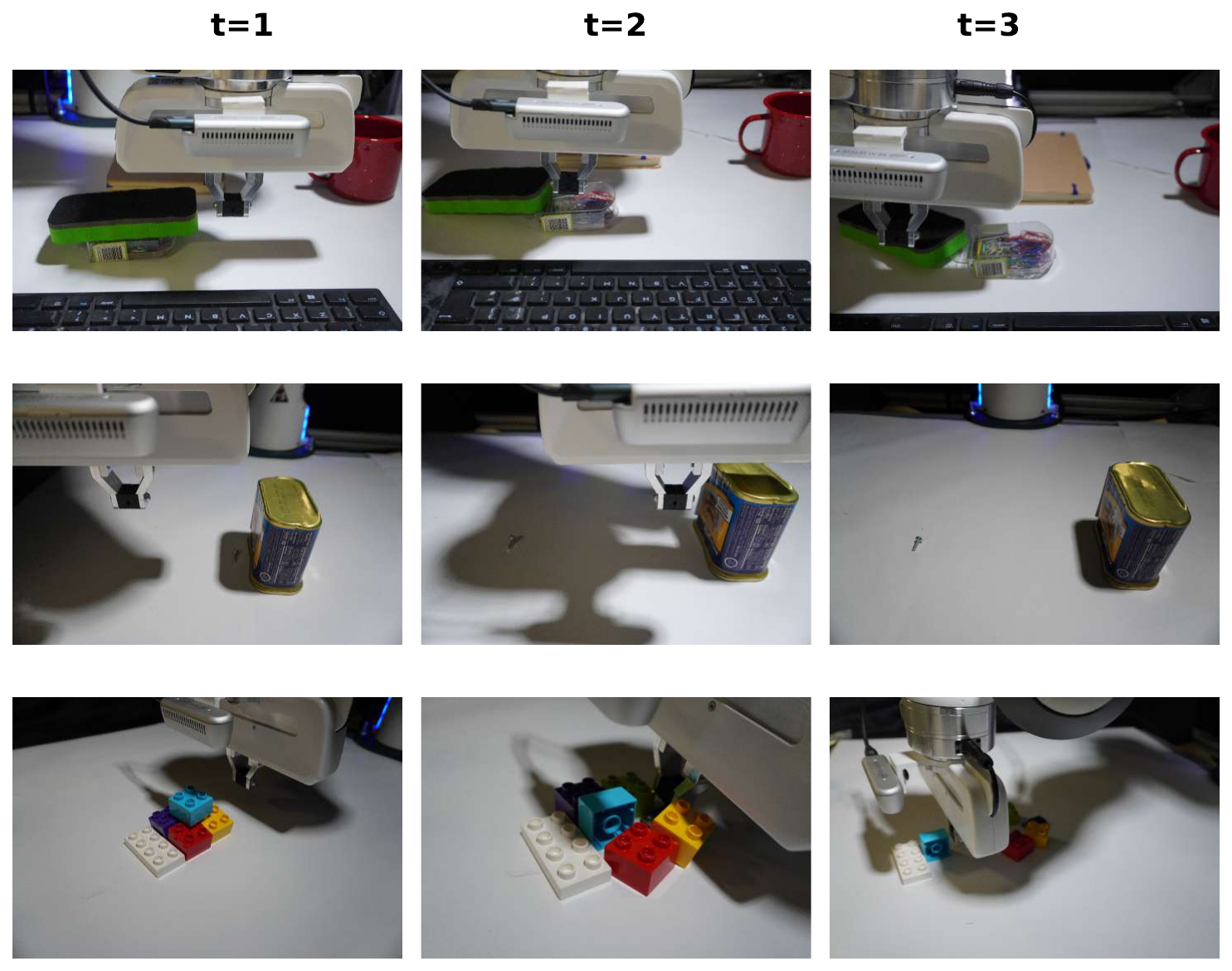}
    \caption{Robot workspace during pushing actions. A source of light produces a strong shadow. The top row shows Task I, uncovering a box of paper clips under the whiteboard cleaner, the middle row shows Task II pushing an aluminium tin, revealing a small screw, and the bottom row shows Task III a clutter of Lego blocks.}
    \vspace{-0.5cm}
    \label{fig:enter-label}
\end{figure}

\begin{figure*}[tb!]
    \centering
    \begin{subfigure}[b]{0.24\textwidth}
        \centering
        \includegraphics[width=\linewidth,trim={0cm 0cm 4cm 0cm},clip]{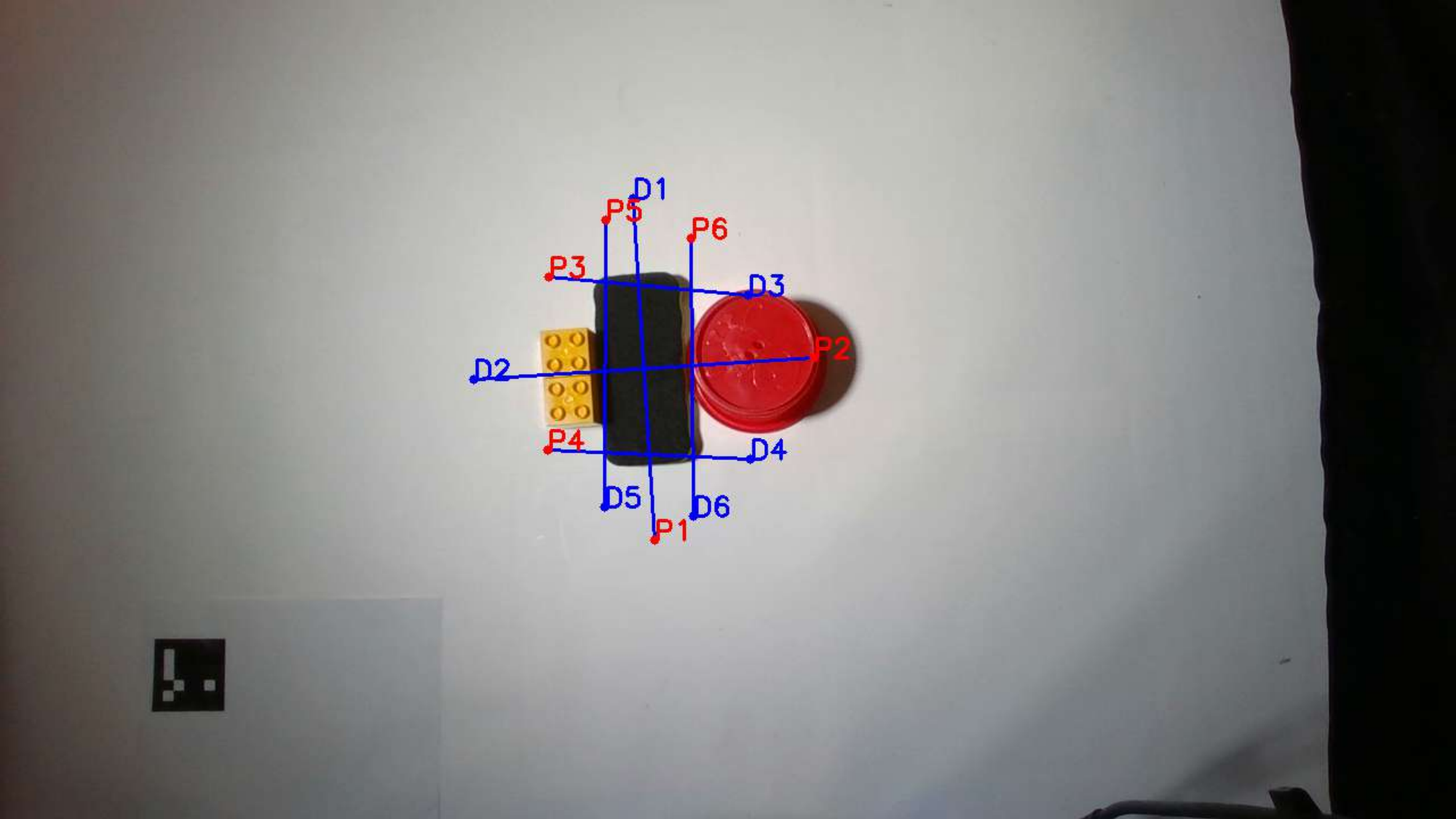}
        \caption{$t=1$}
        \label{fig:scene_7t1}
    \end{subfigure}
    \hfill
    \begin{subfigure}[b]{0.24\textwidth}
        \centering
        \includegraphics[width=\linewidth,trim={0cm 0cm 4cm 0cm},clip]{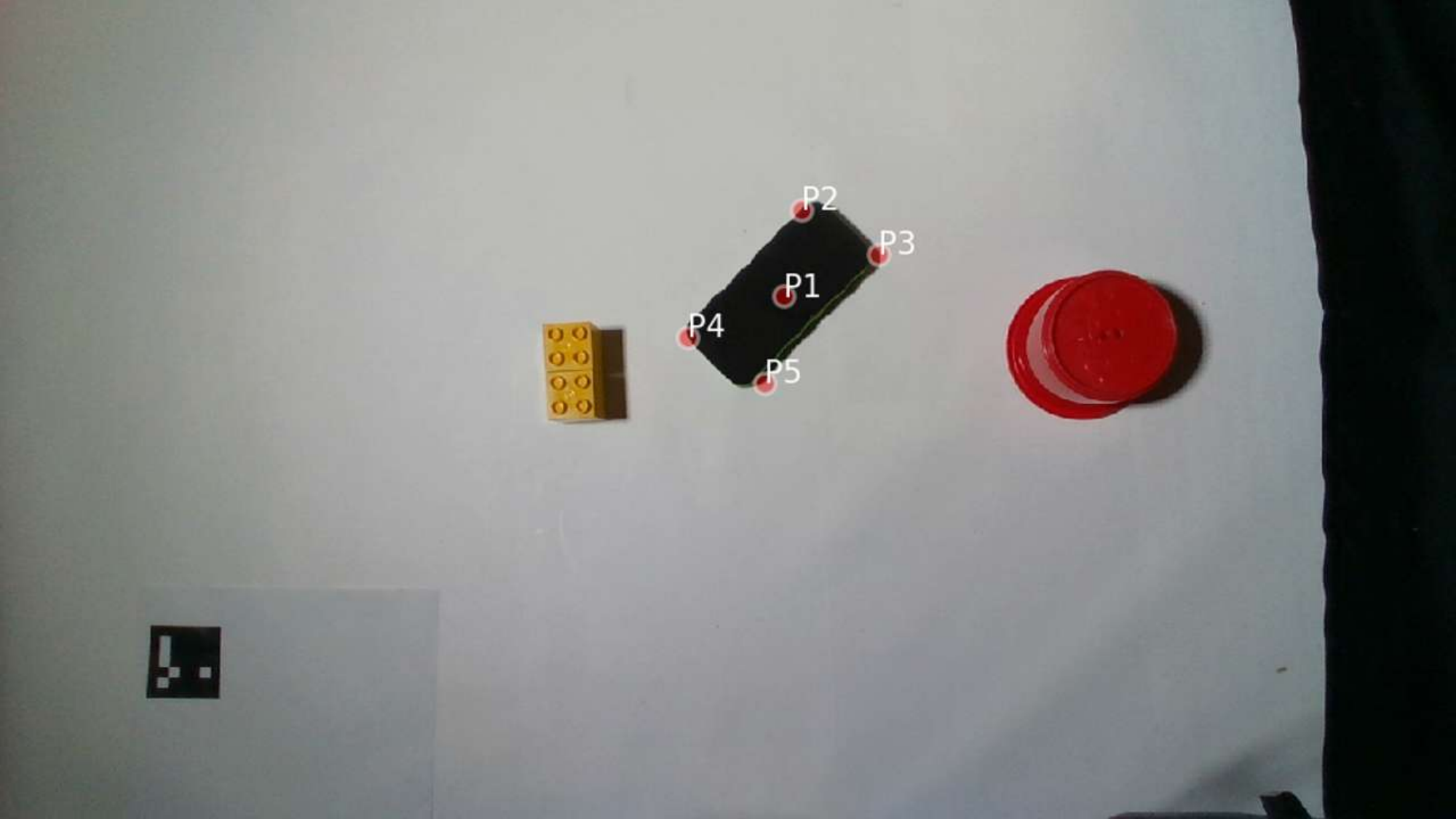}
        \caption{$t=2$}
        \label{fig:scene_7t2}
    \end{subfigure}
    \hfill
    \begin{subfigure}[b]{0.24\textwidth}
        \centering
        \includegraphics[width=\linewidth,trim={0cm 0cm 4cm 0cm},clip]{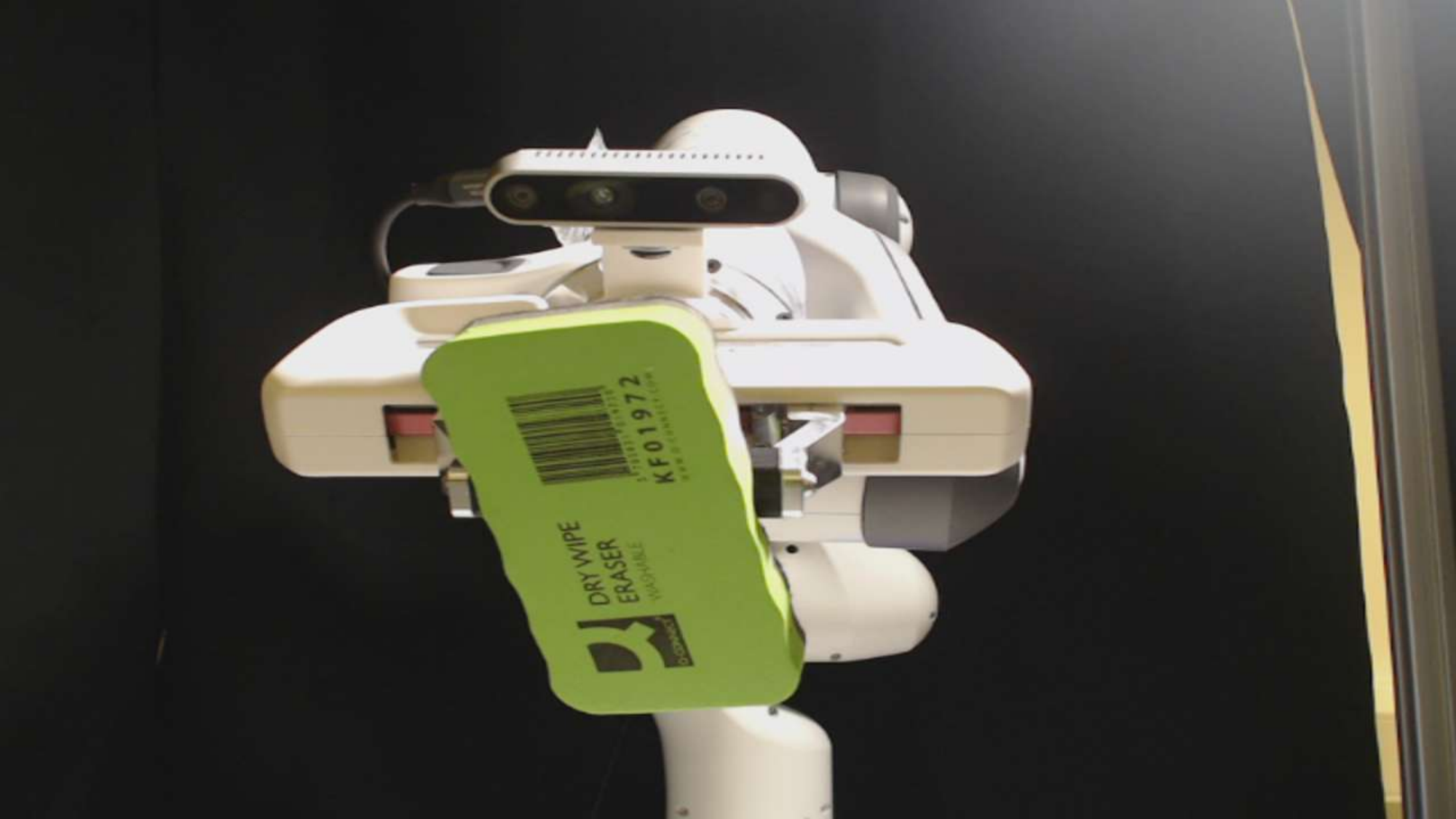}
        \caption{$t=3$}
        \label{fig:scene_7t3}
    \end{subfigure}
    \hfill
    \begin{subfigure}[b]{0.24\textwidth}
        \centering
        \includegraphics[width=\linewidth,trim={1cm 0.1cm 4cm 4.5cm},clip]{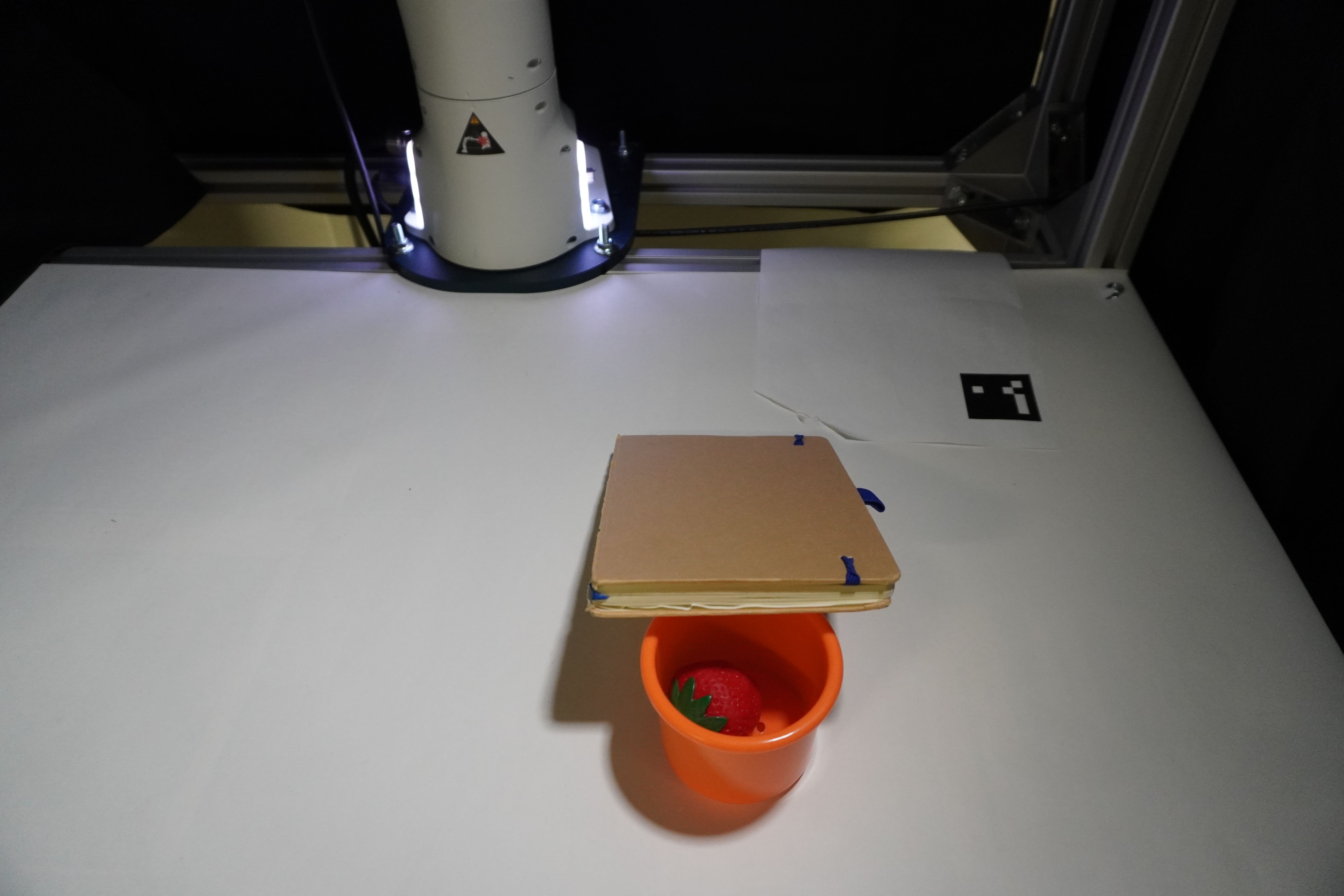}
        \caption{Task VIII}
        \label{fig:scene_7t4}
    \end{subfigure}    
    
    \vspace{-0.1cm} 
    \caption{Task VII being successfully performed at various time steps (query: `What language is the text written in on the black eraser?'): (a) $t=1$: \{$x_1, o_1$\} with pushing action proposed, $EO_P$, (b) $t = 2$: \{$x_2, o_2$\} after pushing, red and yellow objects are separated from the eraser. Grasping key points are now proposed $EO_G$, (c) $t=2$: Cam2 view of eraser that is lifted and moved by the robot in front of the Cam2, (d) Task VIII: brown book partially occludes the view into the cup.}
    \label{fig:task_7_images}
\end{figure*}

Figures \ref{fig:enter-label} and \ref{fig:task_7_images} show Task I-III, VII, and VIII

\subsection{Tasks in Designed Experiments}
\label{sec:tasks_prompts}

\begin{figure*}[t!]
    \centering
    \begin{tabular}{cccc}
        \includegraphics[width=0.22\linewidth]{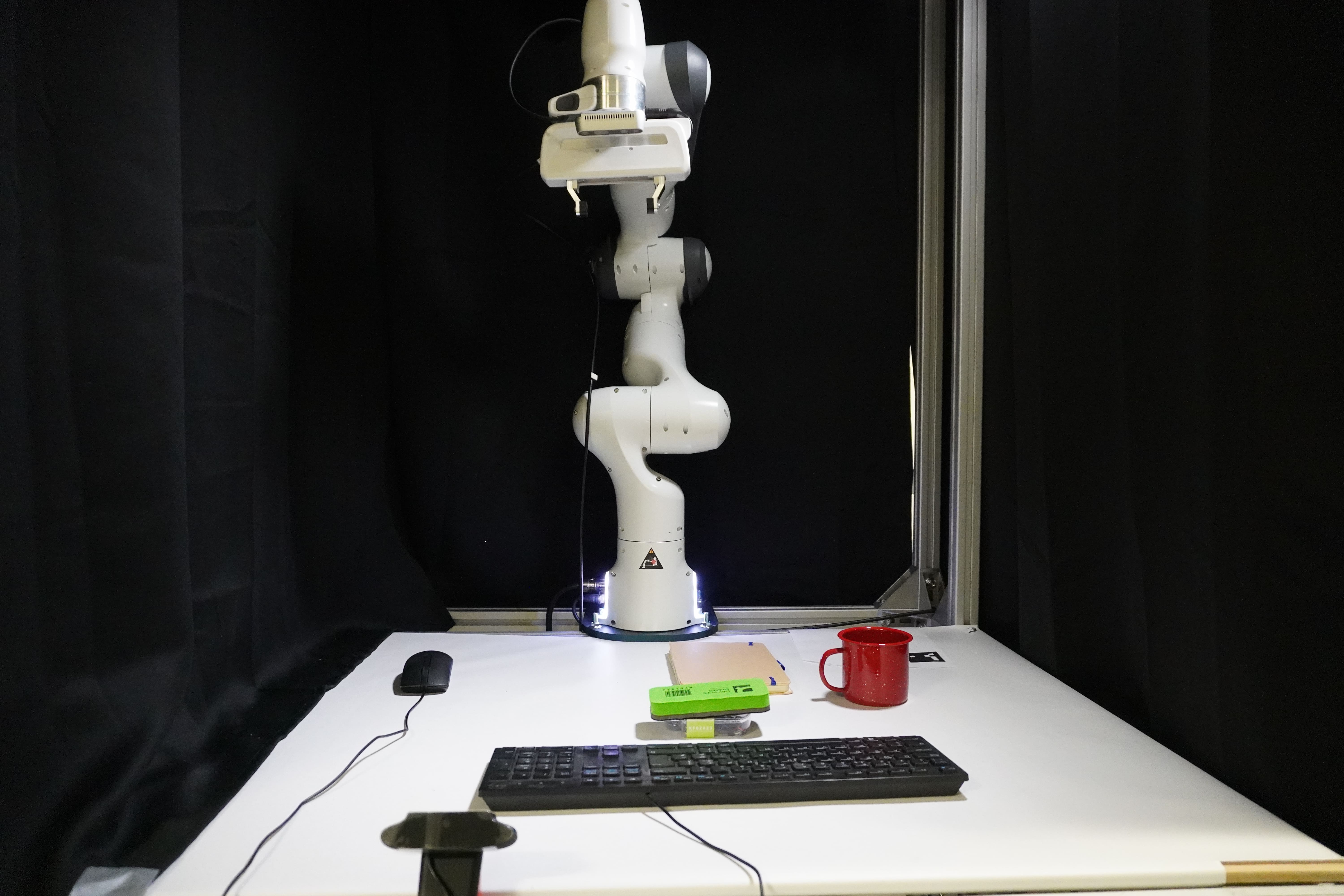} &
        \includegraphics[width=0.22\linewidth]{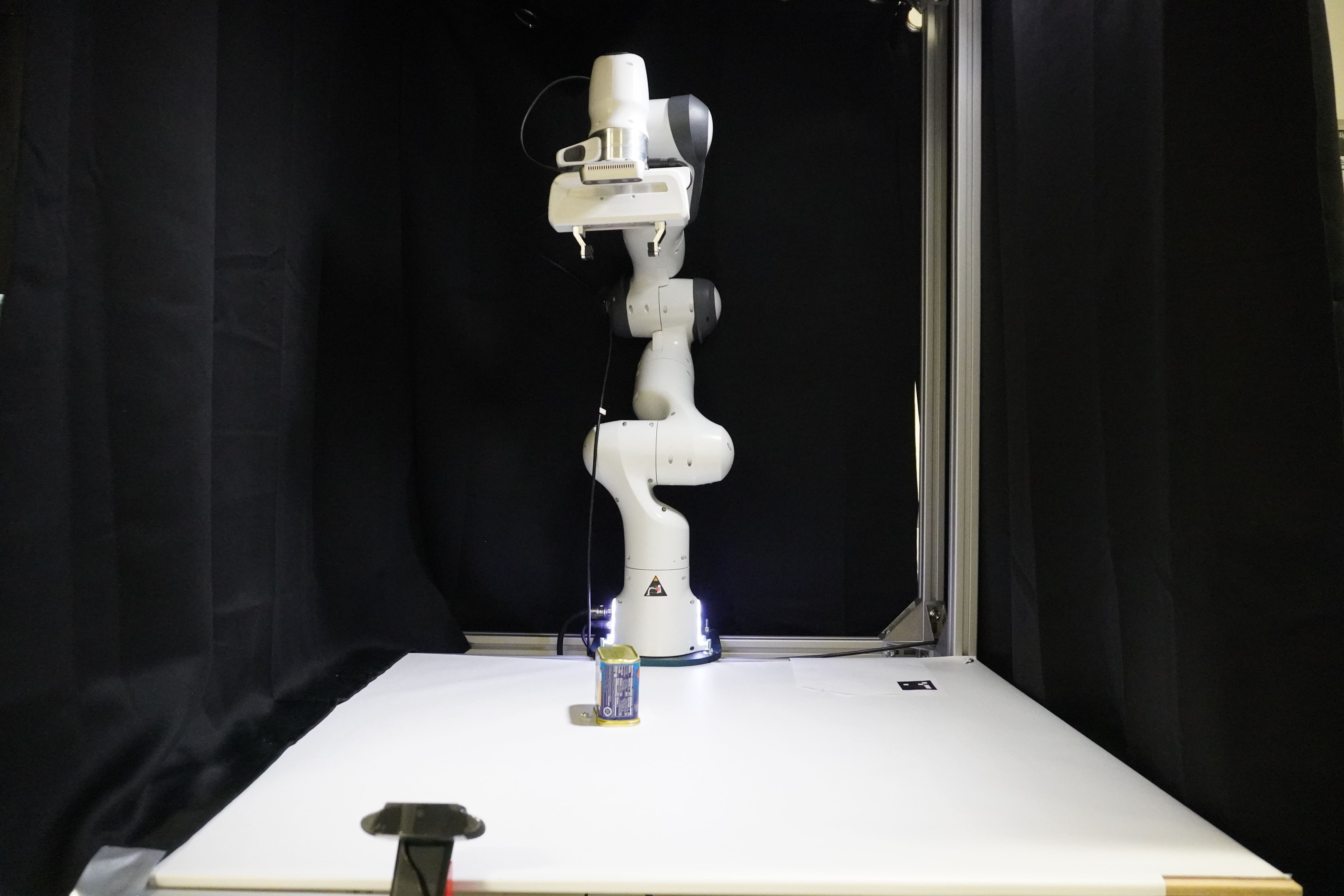} &
        \includegraphics[width=0.22\linewidth]{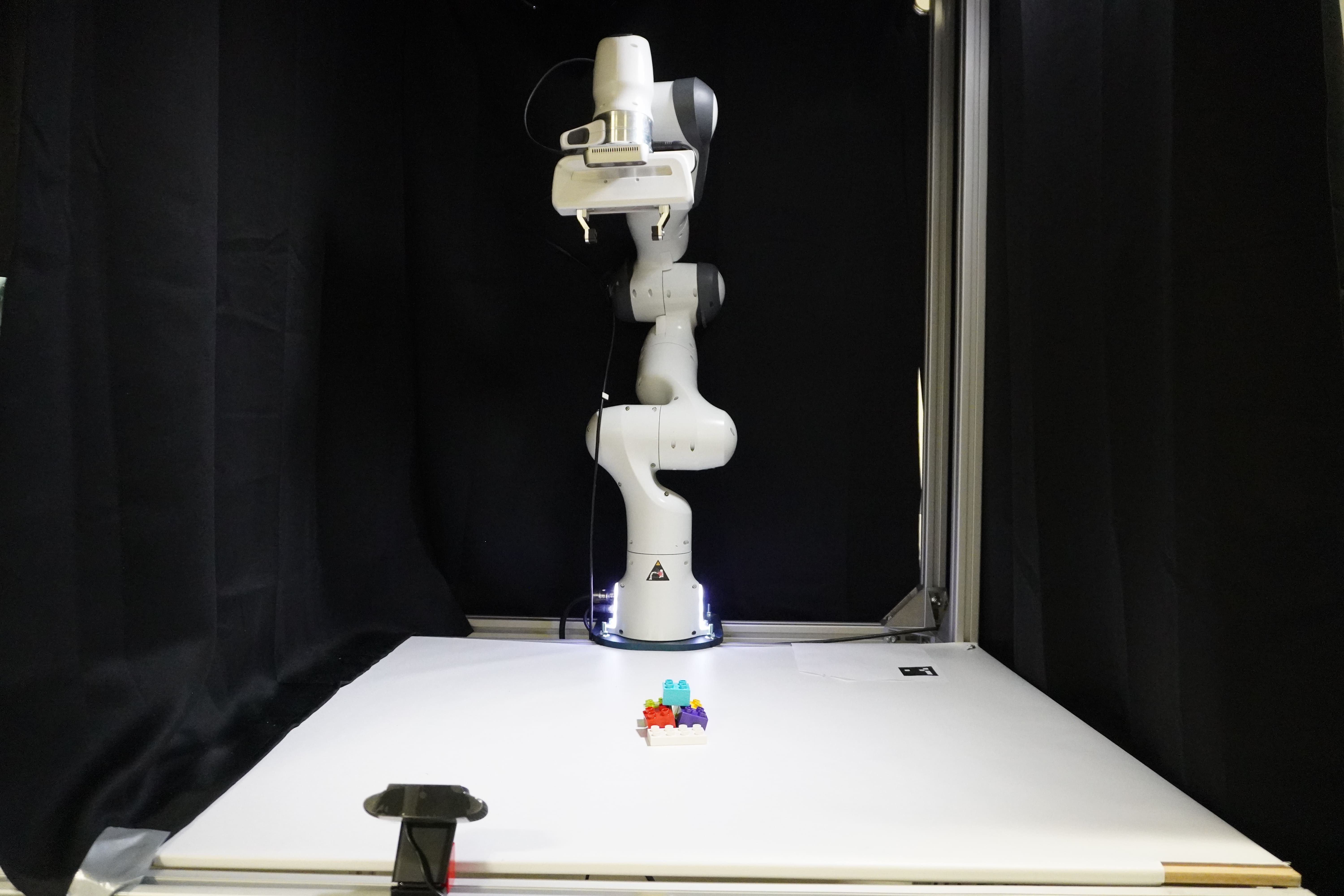} &
        \includegraphics[width=0.22\linewidth]{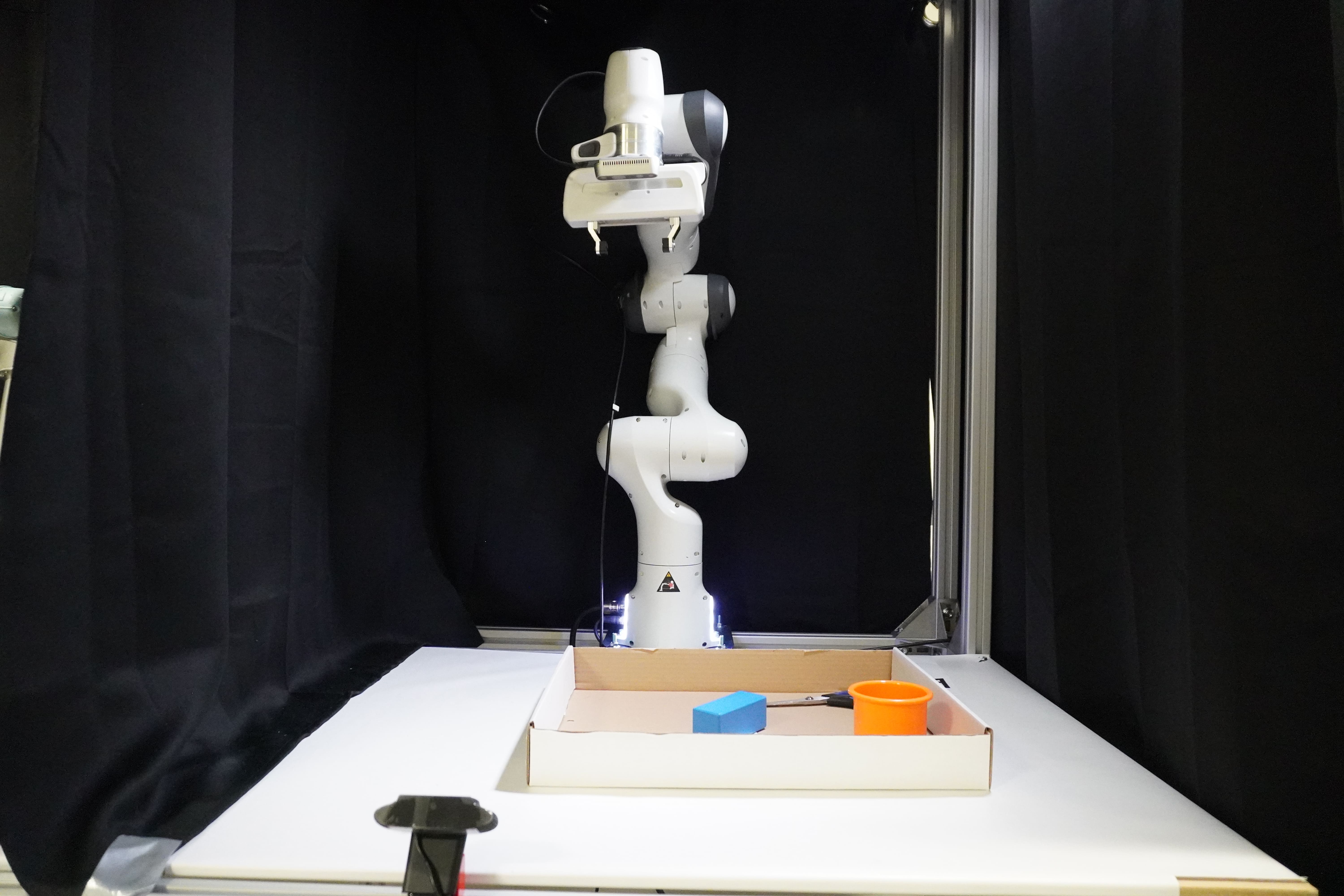} \\
        
        \includegraphics[width=0.22\linewidth]{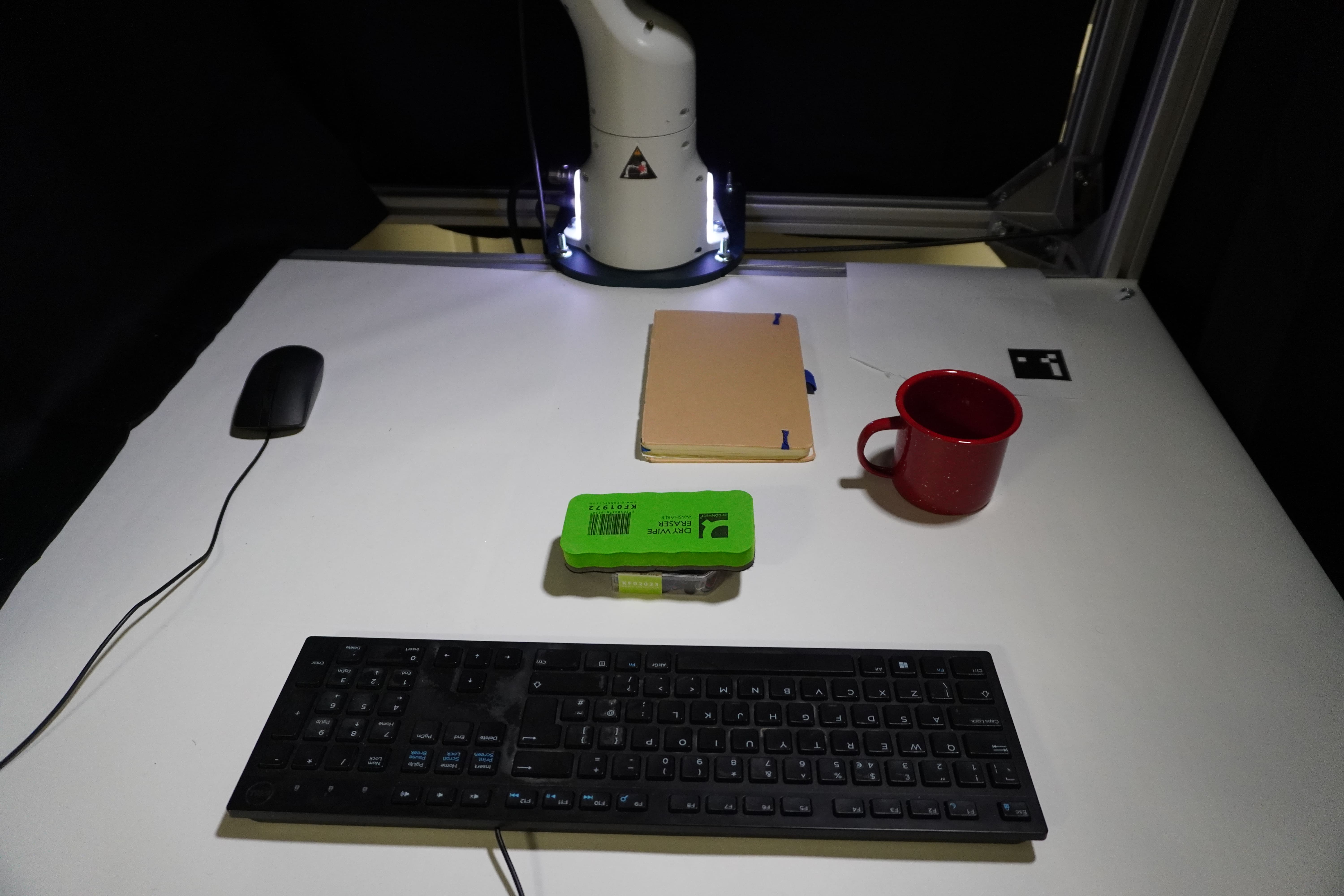} &
        \includegraphics[width=0.22\linewidth]{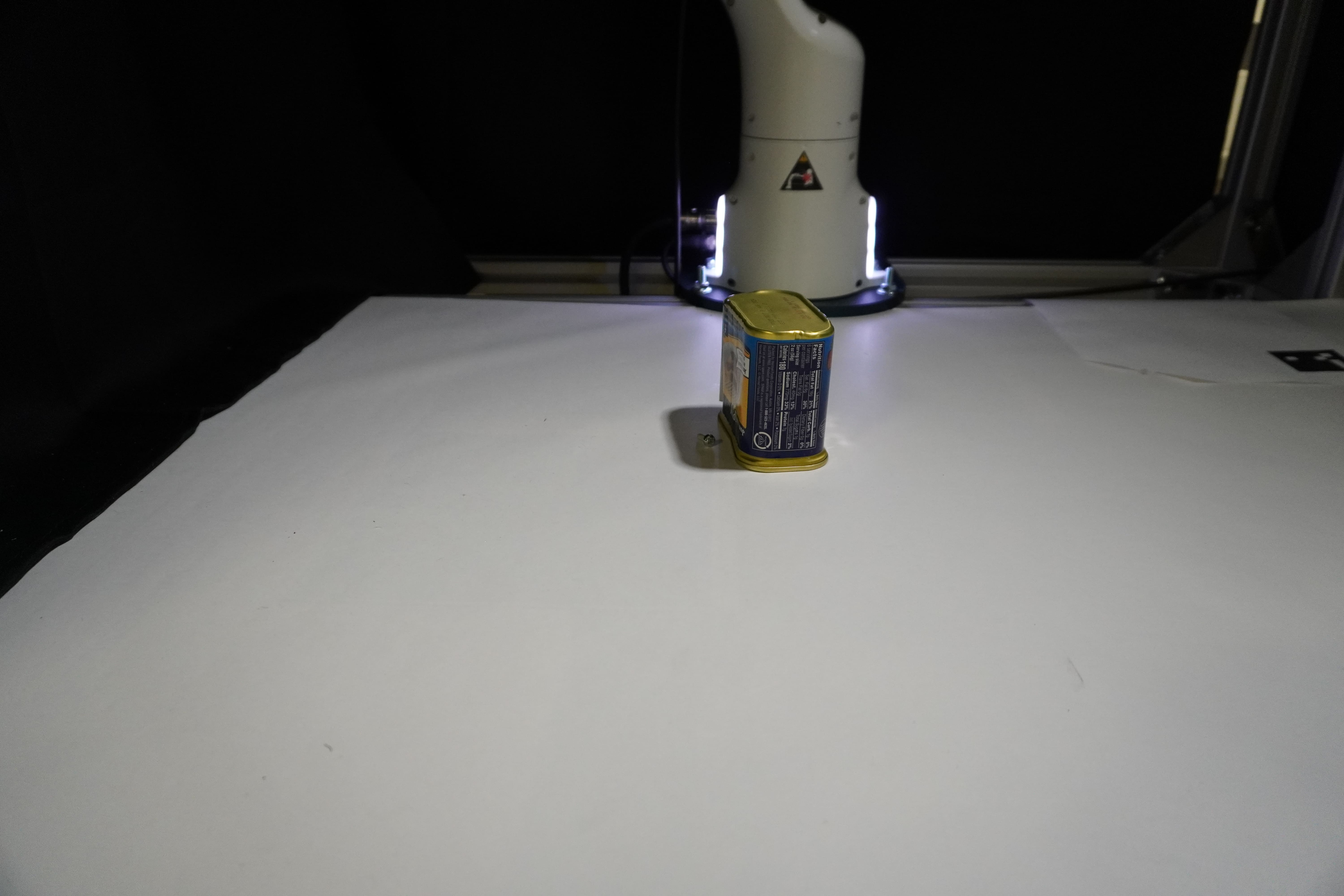} &
        \includegraphics[width=0.22\linewidth]{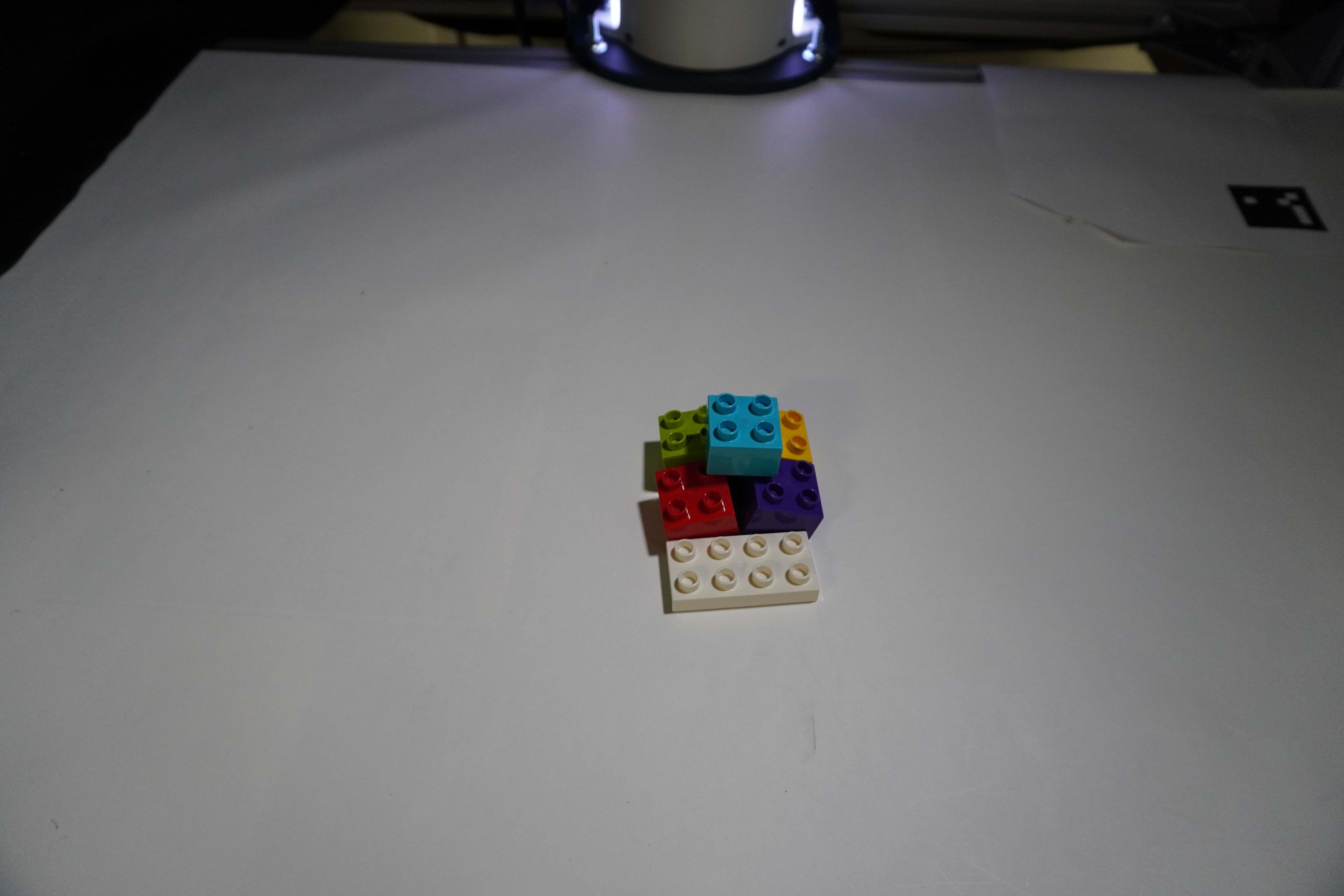} &
        \includegraphics[width=0.22\linewidth]{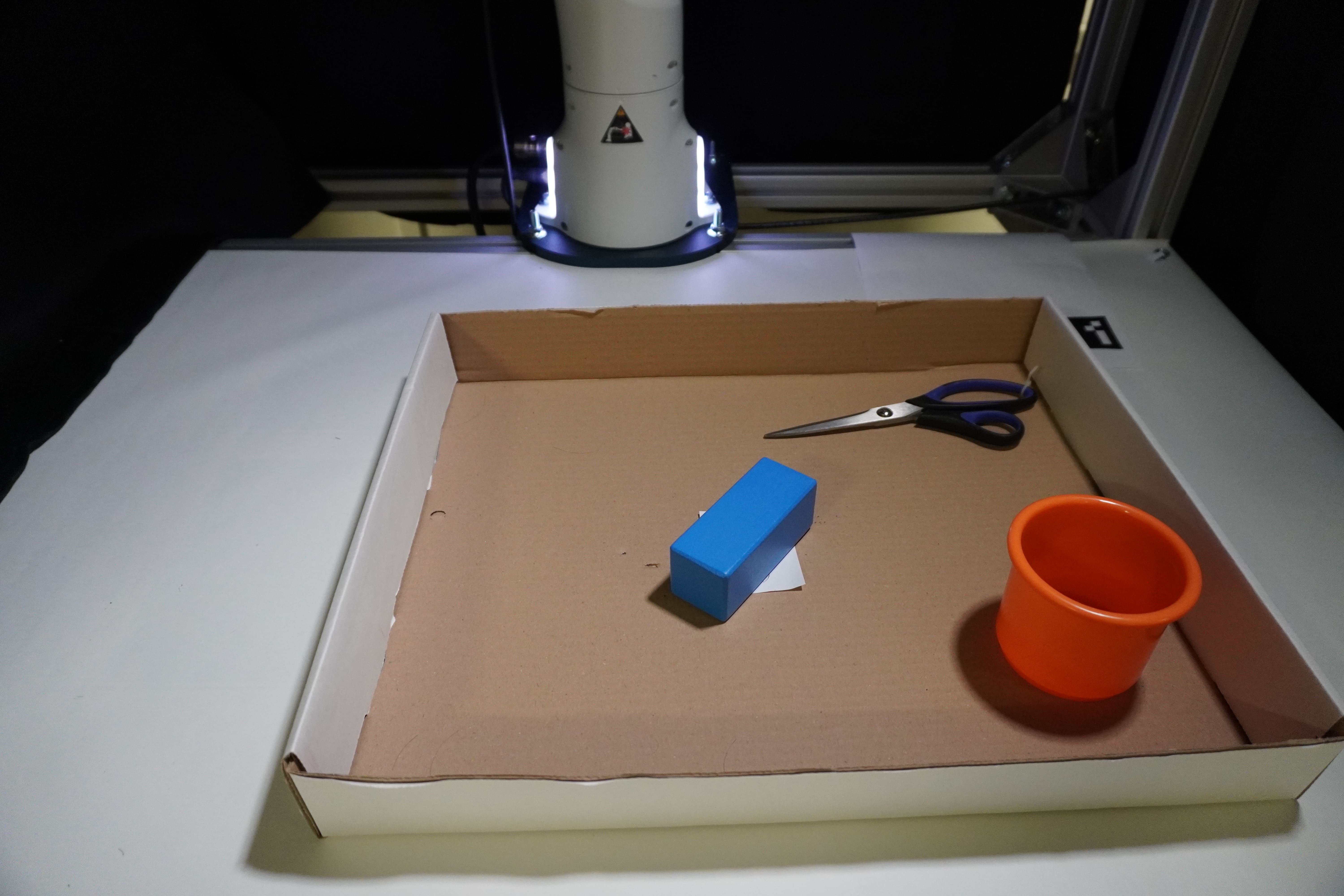} \\
        
        \includegraphics[width=0.22\linewidth]{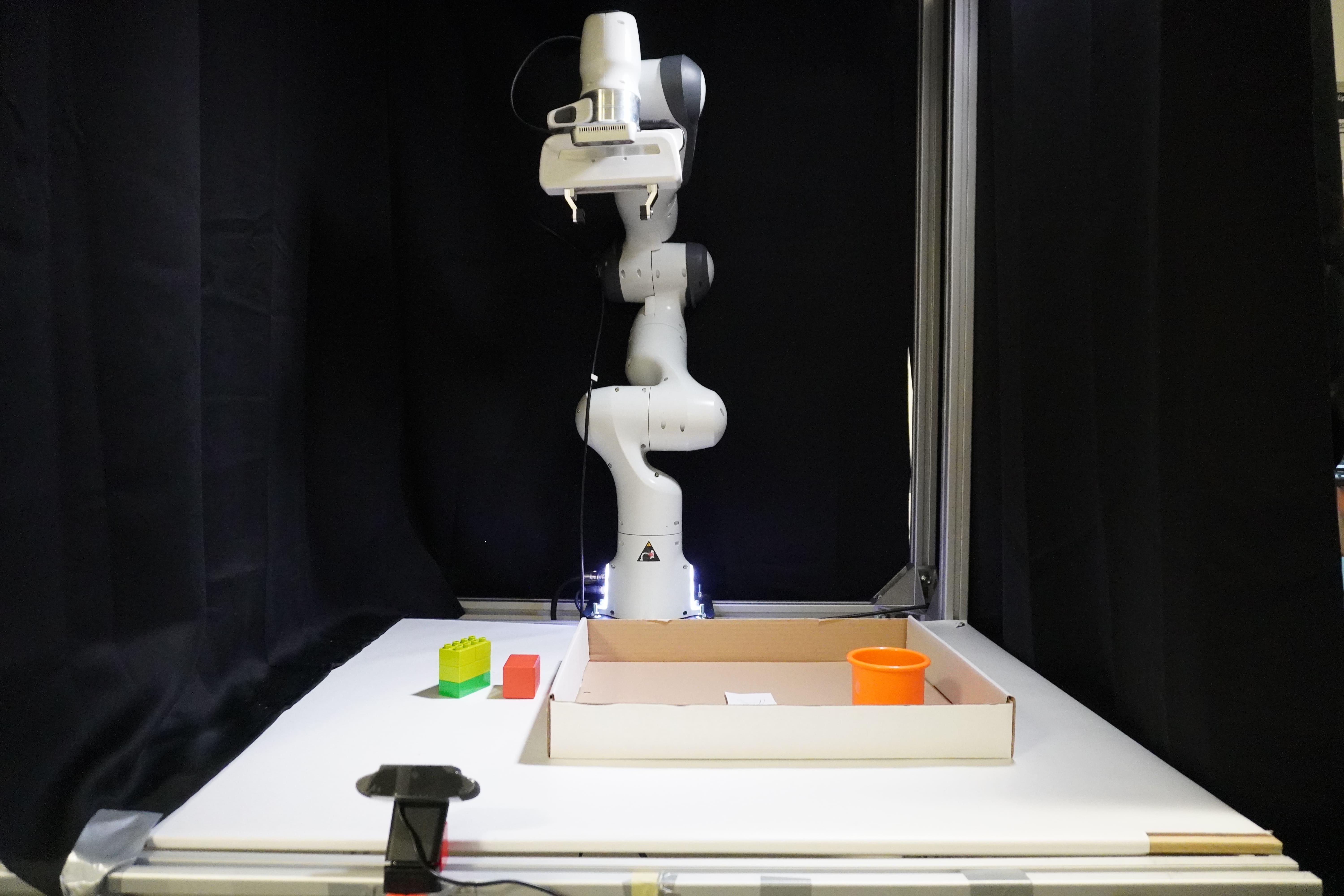} &
        \includegraphics[width=0.22\linewidth]{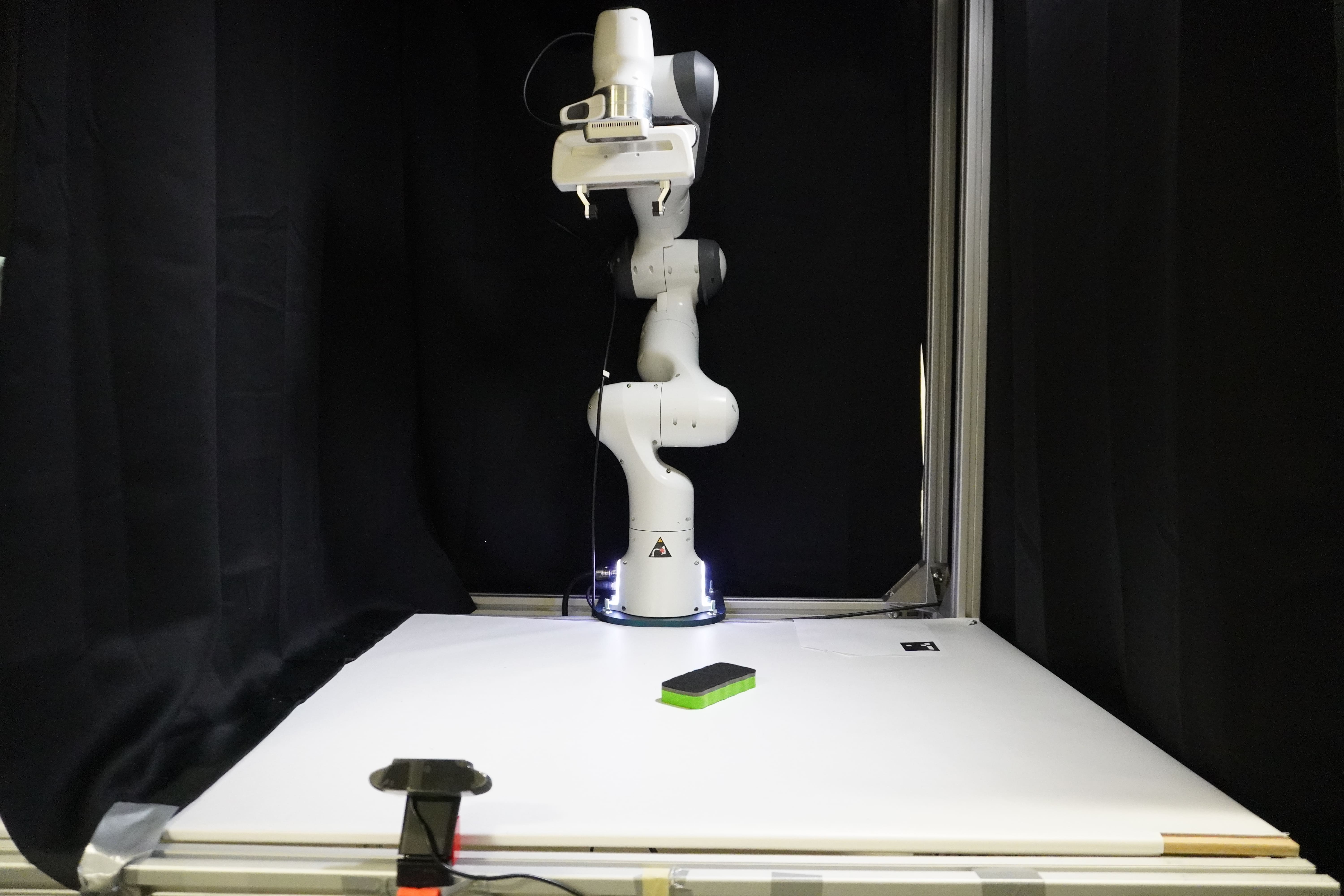} &
        \includegraphics[width=0.22\linewidth]{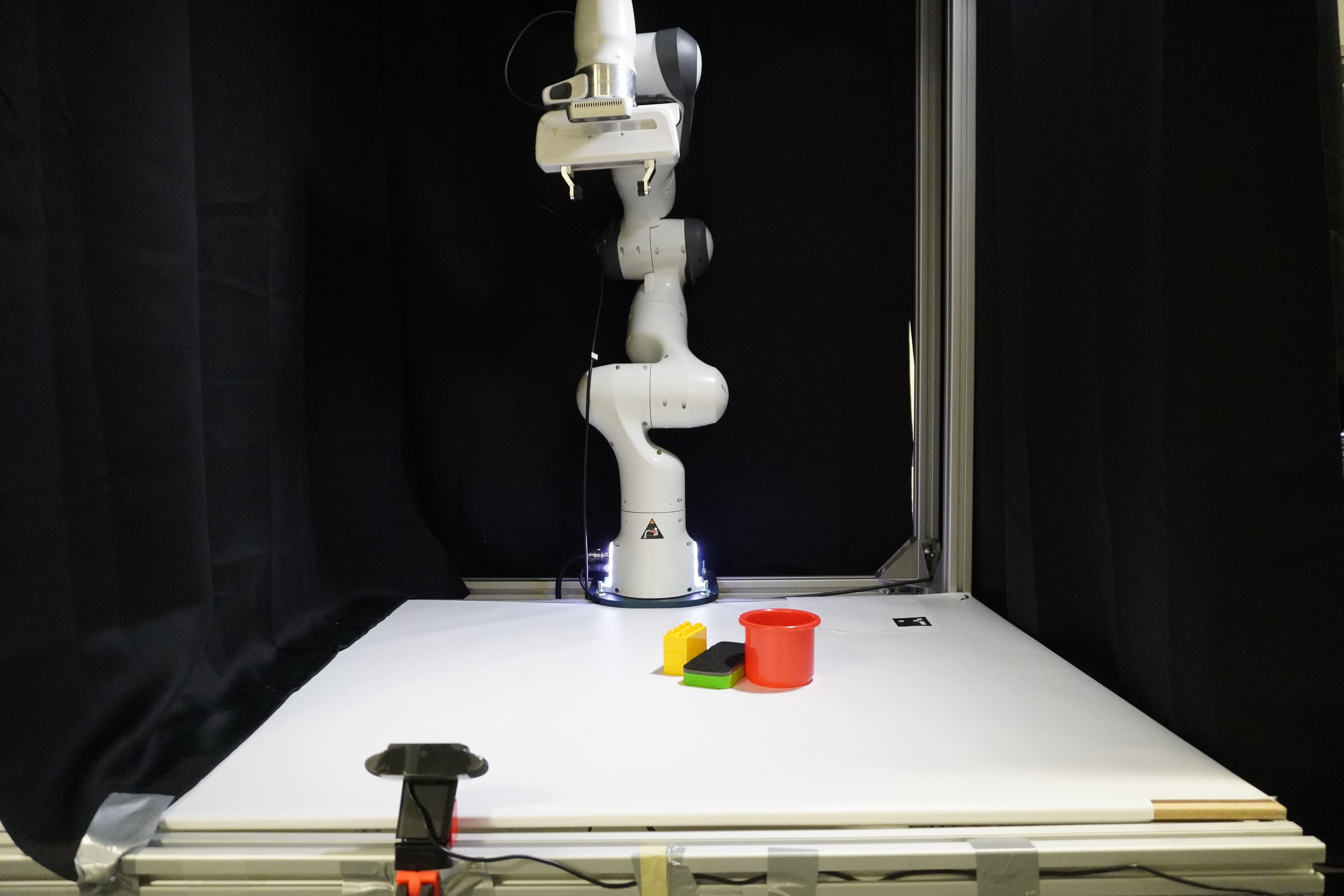} &
        \includegraphics[width=0.22\linewidth]{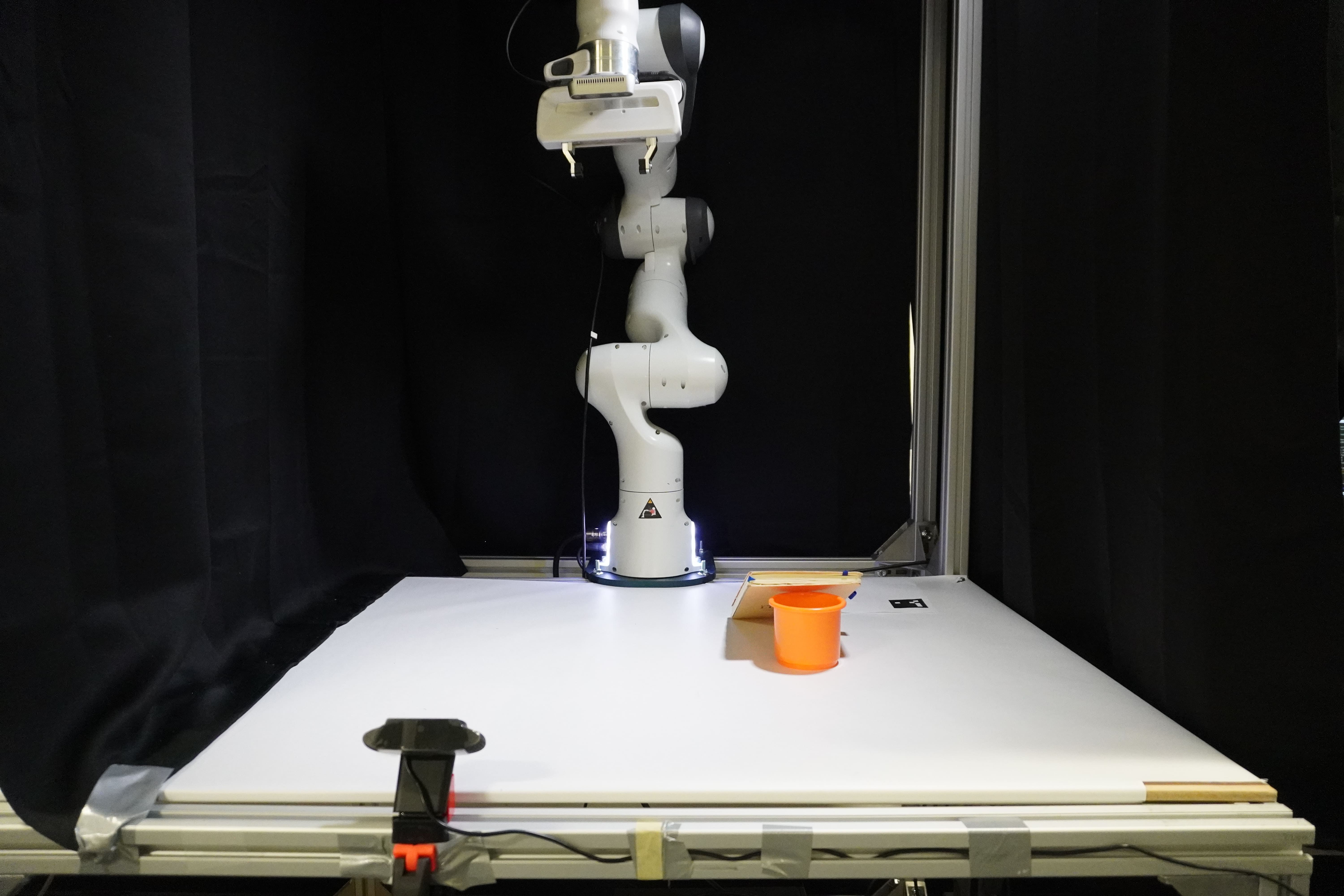} \\
        
        \includegraphics[width=0.22\linewidth]{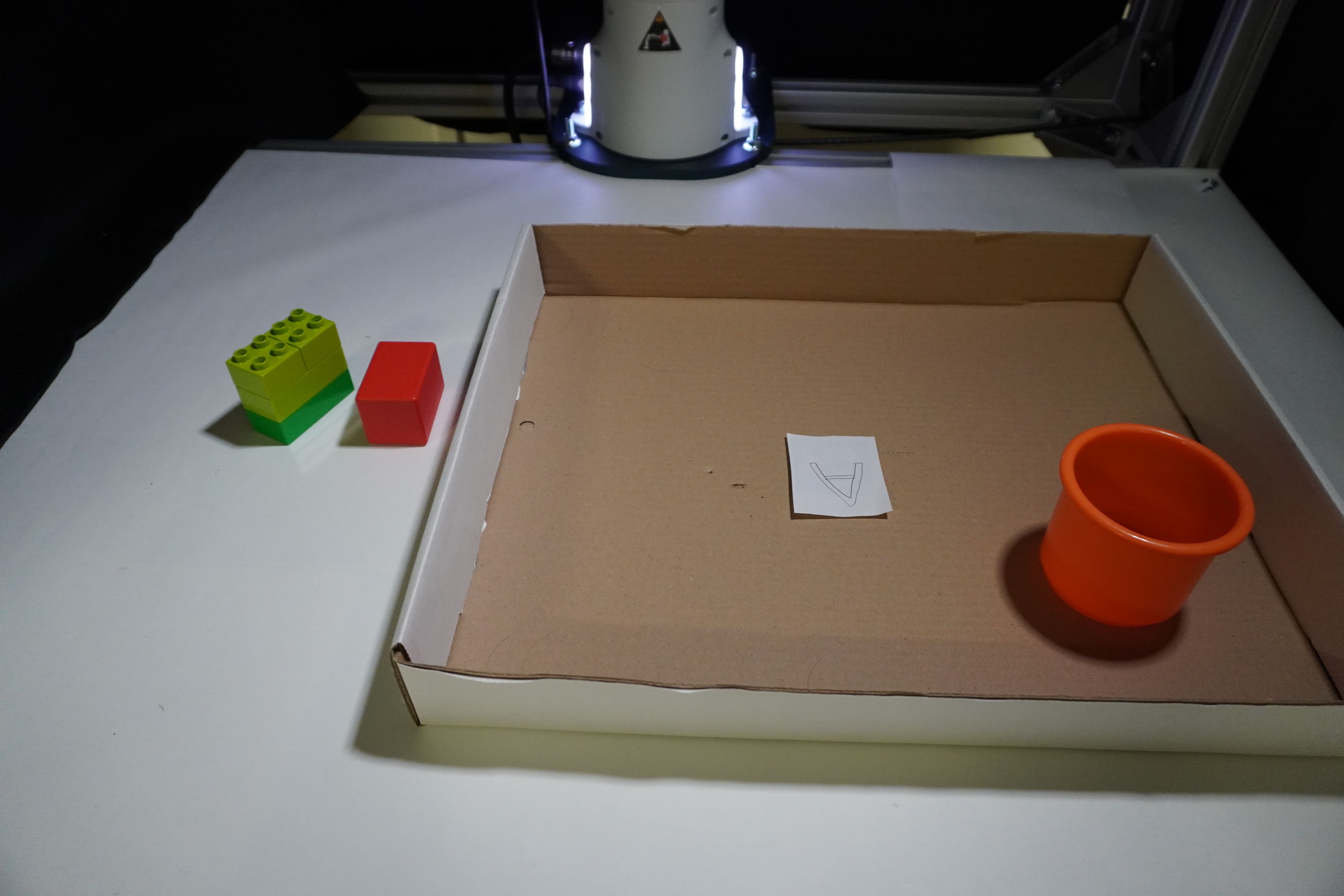} &
        \includegraphics[width=0.22\linewidth]{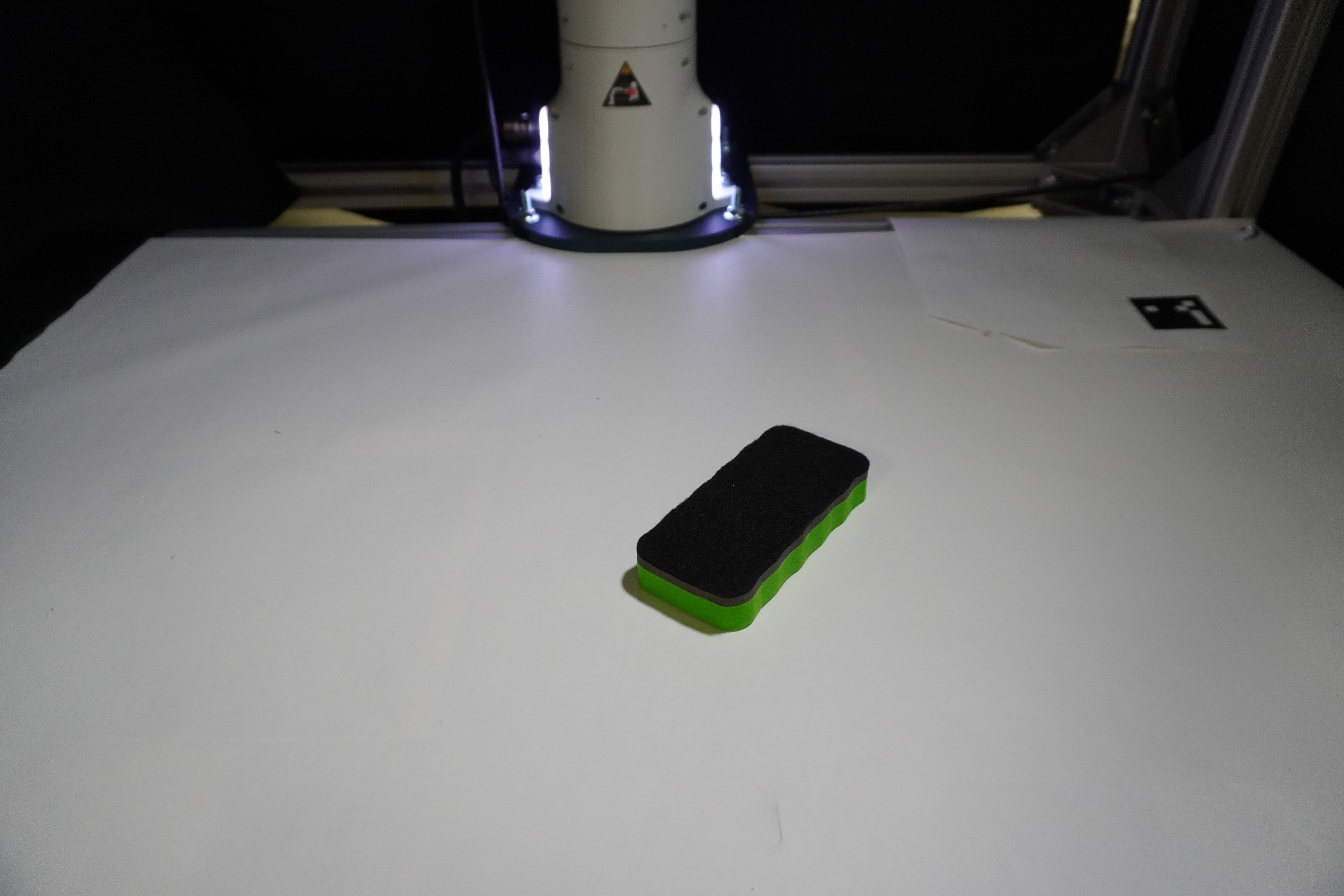} &
        \includegraphics[width=0.22\linewidth]{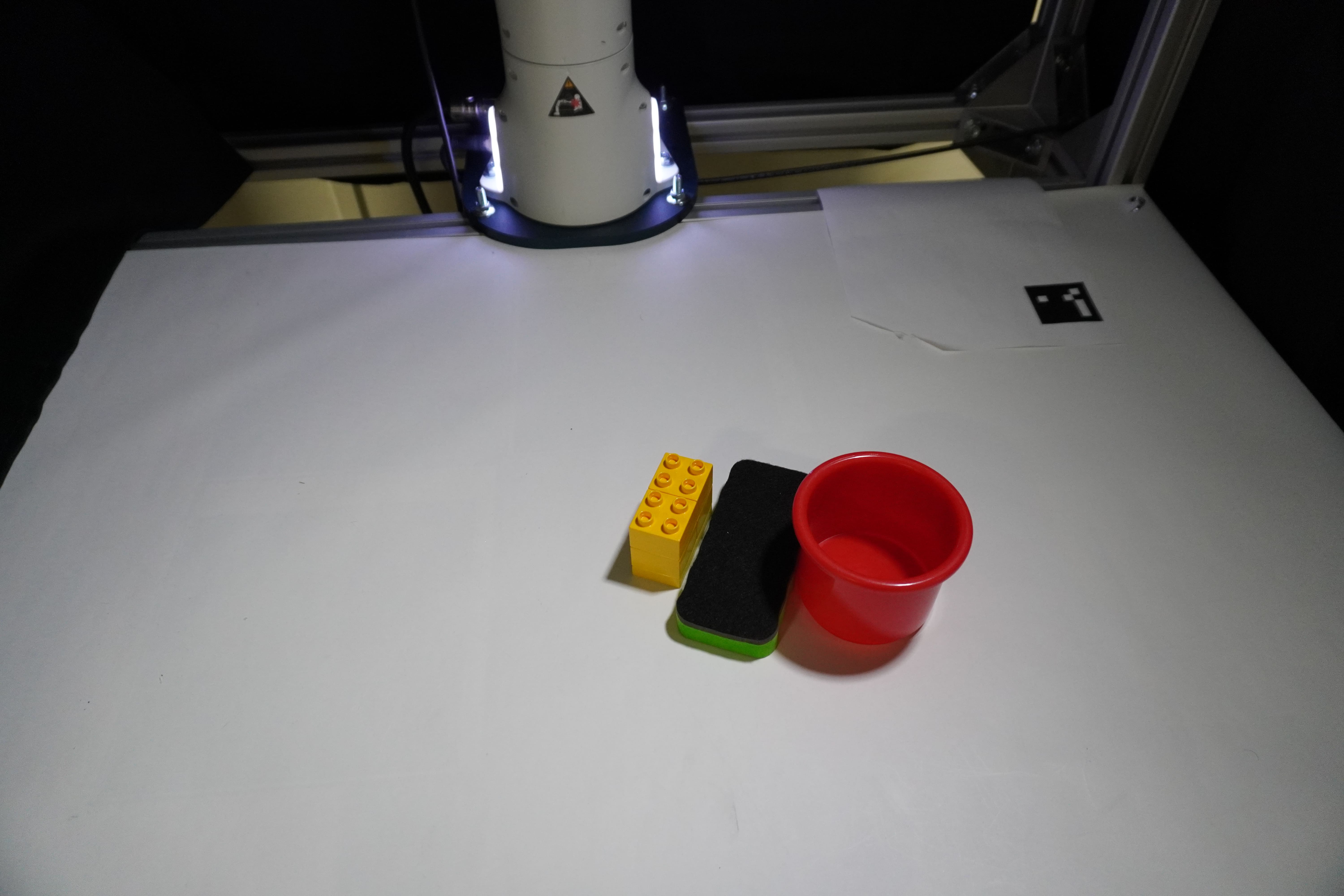} &
        \includegraphics[width=0.22\linewidth]{Figures/rss25/appendix/task_images/task_8/DSC00869.JPG} \\
    \end{tabular}    
    \caption{(From left to right) First and Third row: Task I to VIII. Second and Fourth row: the robot's workspace in Task I to VIII. Task I: the robot needs to push the eraser to see the box of paper clips underneath. Task II: The robot needs to pick up the aluminium tin and place it elsewhere to see the screw in the shadow. Task III: Push the Lego blocks to see the pen-drive underneath. Task IV: Pick up the blue block and place it elsewhere to see the text written on the paper under it. Task V: Identify the red block and green Lego structure on the table. Arrange them inside the cardboard box while maintaining their original spatial positions relative to one another. Task VI: grasp the black eraser and bring it in front of the secondary camera to see what is written on the side under it. Task VII: similar to Task VI but eraser is between two objects. Hence ZS-IP requires first pushing the eraser and then grasping to complete the task. Task VIII: ZS-IP should see what is inside the cup covered by the brown book. }
    \label{fig:tasks}
\end{figure*}

Fig.~\ref{fig:tasks} presents the tasks used in the ZS-IP evaluation. 

\begin{itemize}
    \item \textbf{Task I:} The robot pushes the eraser to reveal the box of paper clips beneath it.
    \item \textbf{Task II:} The robot picks up the aluminium tin and relocates it to expose the screw hidden in the shadow.
    \item \textbf{Task III:} The robot pushes the Lego blocks to uncover the pen drive underneath.
    \item \textbf{Task IV:} The robot picks up the blue block and moves it to reveal the text written on the paper beneath it.
    \item \textbf{Task V:} The robot identifies the red block and green Lego structure on the table, arranging them inside the cardboard box while preserving their original spatial relationships.
    \item \textbf{Task VI:} The robot grasps the black eraser and brings it in front of the secondary camera to examine the text written on its underside.
    \item \textbf{Task VII:} Similar to Task VI, but with the eraser positioned between two objects, requiring ZS-IP to first push the eraser before grasping it to complete the task.
    \item \textbf{Task VIII:} The robot determines the contents of a cup covered by a brown book.
\end{itemize}

Task queries and prompts are detailed in Section~\ref{sec:tasks_prompts}.




\subsection{Ablation studies}
\label{appendix:ablation}

Table \ref{tab:in_depth_results_PIVOT} shows that ZS-IP outperforms PIVOT with and without interactive perception. This is attributed to the improvement in VLM visual reasoning capability induced by the Enhanced Observation (EO).

Tables \ref{tab:gemini_claude} and \ref{tab:in_depth_results} compare the performance of GPT-4o with other VLMs and varying VLM version within GPT. We only compare for one task each from "grasping", "lifting", and "observe" to gain an understanding of the output and reasoning capabilites of the various models.

\begin{table*}[t!]
\centering
\caption{Comparing ZS-IP with different features and PIVOT. The performance Metrics include SR (Success Rate), TL (Total Length) TLP (Total Length Successful), PE (Position Error) and OSR (Oracle Success Rate).}
\label{tab:in_depth_results_PIVOT}
\resizebox{\linewidth}{!}{
\begin{tabular}{|l|c|c|c|c||c|c|c|c||c|c|c|c|}
\hline
\multirow{2}{*}{\textbf{Method}} & \multicolumn{4}{c||}{\textbf{Task IV}} & \multicolumn{4}{c|}{\textbf{Task VI}} & \multicolumn{4}{c|}{\textbf{Task VIII}} \\ \cline{2-13}
 & \textbf{SR $\uparrow$} &\textbf{TL, TLS $\downarrow m$} & \textbf{PE} $\downarrow m$ & \textbf{OSR $\uparrow$} & \textbf{SR $\uparrow$} & \textbf{TL,TLS $\downarrow m$} & \textbf{PE $\downarrow m$} & \textbf{OSR} $\uparrow$ & \textbf{SR $\uparrow$} & \textbf{TL,TLS $\downarrow m$} & \textbf{PE $\downarrow m$} & \textbf{OSR} $\uparrow$ \\ \hline
 
PIVOT w interact perception             & 0.0         & \{7.71, -\}           & 0.55         & 14         & 0.7         & \{4.86, 3.81\}         & \textbf{0.541}         & \textbf{7}  & 0.0         & \{5.78, -\}         & \textbf{0.59}         & 1        \\ \hline

PIVOT w active perception  & -         & -           & -         & -         & -        & - & -         & -         & 0.3         & \{1.06, 0.254\}         & \textbf{0.428}         & 0        \\ \hline

ZS-IP             & \textbf{1.0}         & \textbf{\{1.83, 1.83\}}          & \textbf{0.64}        & 1         & \textbf{0.8}         & \textbf{\{2.16, 2.11\}}         & 0.59         & 3       & 0.7         & \textbf{\{1.47, 1.56\}}         & 0.64         & \textbf{4}        \\ \hline

\end{tabular}
}
\end{table*}

\begin{table*}[h!]
\centering
\caption{ZS-IP performed on Gemini 2.0 Flash and Claude 3.5 Sonnet}
\label{tab:gemini_claude}
\resizebox{\textwidth}{!}{ 
\footnotesize
\begin{tabular}{lcccccccc}
\toprule
\multirow{4}{*}{\textbf{Method}} & 
\multicolumn{3}{c}{\textbf{Pushing}} & 
\multicolumn{2}{c}{\textbf{Grasping}} & 
\multicolumn{2}{c}{\textbf{Lifting to Investigate}} &
\multicolumn{1}{c}{\textbf{Observe}}\\ 

\cmidrule(lr){2-4} \cmidrule(lr){5-6} \cmidrule(lr){7-8} \cmidrule(lr){8-9} 
 & \textbf{Task I} & \textbf{Task II} & \textbf{Task III} & 
\textbf{Task IV} & \textbf{Task V} & 
\textbf{Task VI} & \textbf{Task VII} & \textbf{Task VIII} \\
\midrule
ZS-IP w GPT-4o (Ours) & 0.9 & 0.9 & 0.6 & 1.0 & 0.8 & 0.8 & 0.2 & 0.7\\
ZS-IP w Gemini 2.0 Flash & 0.5 & 0.6 & - & 0.9 & - & - & - & -\\
ZS-IP w Claude 3.5 Sonnet & 0.8 & 1.0 & - & 1.0 & - & - & - & -\\
\bottomrule
\end{tabular}
}
\label{tab:ip_vlm_3vlms}
\end{table*}

\begin{table}[t!]
\centering
\caption{Comparison of GPT-4o with Other Models}
\label{tab:in_depth_results_comparison_gpt}
\resizebox{0.3\linewidth}{!}{
\begin{tabular}{|l|c||c|}
\hline
\multirow{2}{*}{\textbf{Method}} & \multicolumn{1}{c||}{\textbf{Task III}} \\
 & \textbf{SR $\uparrow$} \\ \hline
GPT-4o              & 0.6       \\ \hline
GPT-4 Turbo         & 0.0        \\ \hline
\end{tabular}
}
\end{table}

\subsection{Prompts}

The Perception Analyser processes the scene and generates an output guiding ZS-IP to the next step. Appendix Figure~\ref{appendix:pa_explain} illustrates the base prompt. In the first iteration, the Perception Analyser operates without memory, while from the second iteration onward, memory is introduced in the form of prior actions, thoughts, and images. The template is presented in Appendix Figure~\ref{appendix:pa_memory}, and an example output is shown in Appendix Figure~\ref{appendix:pa_output}.


\begin{figure*}[t!]
\begin{tcolorbox}[title=Perception Analyser]
You are an intelligent embodied agent that can thoroughly perform object detection and action planning. Your job is to analyse the provided image and precisely answer the input task or text question, selecting and justifying an appropriate action if required. The image is captured by a camera mounted on a Franka Panda robotic manipulator.

\#\#\# Instructions:

    1. Understand the Question: 
    
        - Identify the Goal Object mentioned or implied in the question.
        
        - If no Goal Object is specified, focus on the general intent of the question.

    2. Analyze the Image:
    
        - Detect and list all visible objects in the image.
        
        - Determine if the Goal Object is fully visible, partially obscured, or completely hidden.

    3. Take Action:
    
        - If the Goal Object is obscured:
        
            \hspace{1cm}- Identify the object obstructing it the most (the Target Object).
            
            \hspace{1cm}- Analyse the scene, focusing on the key details relevant to answering the question. Think about one action or series of actions that need to be taken to complete the task
            
        - If the Goal Object is visible:
        
            \hspace{1cm}- Answer the question based on its attributes or position.

    4. Decision:
    
        - State if you have conclusively completed the task in question now (Yes/No).
        
        - While some tasks can be completed using only one action, others may require a sequence (e.g., "push" followed by "grasp").
        
\end{tcolorbox}
\caption{Perception Analyser}
\label{appendix:pa_explain}
\end{figure*}

\begin{figure*}[th!]
\begin{tcolorbox}[title=Memory]

You are currently at time step: $\langle \text{time\_step} \rangle$.
        
Your suggestion of the action and the target object in the previous time steps along with their summary and reasoning is given below:

-----

In step $\langle \text{previous\_step} \rangle$:

The action taken was $\langle \text{action} \rangle$. Thought and reasoning for action taken was - $\langle \text{analysis of action taken} \rangle$; thought was $\langle\text{thought process in previous\_step} \rangle$ 
 
Object $\langle \text{action} \rangle$ during scene $\langle \text{previous\_step} \rangle$ was $\langle \text{object in previous\_step} \rangle$

-------

...and so on
(history of all previous steps provided in the above template)

In addition to the above information, you are currently given all the previous RGB images upto time step 0

\end{tcolorbox}
\caption{Memory block of Perception Analyser}
\label{appendix:pa_memory}
\end{figure*}

\begin{figure*}[t]
\begin{tcolorbox}[title=Output]
    - If you have conclusively completed the task in question, provide:
    
    ```
    
        \hspace{1cm}- Answer: [Your Answer]
        
        \hspace{1cm}- Output:stop
    ```
    
   - If you have not conclusively completed the task yet, provide:
   
    ```
    
        \hspace{1cm}- Objects: [List of all detected objects]
        
        \hspace{1cm}- Thought: [Analysis and summary of scene. Explain your thinking about action(s) to take and omit]
        
        \hspace{1cm}- Target Object: [Color followed by name of object to move]
        
        \hspace{1cm}- Analysis of action: [Analysis of scene and reasoning about action to take]
        
        \hspace{1cm}- Action: [Your one suggested action]
        
        \hspace{1cm}- Output:go
        
    ```        
\end{tcolorbox}
\caption{Output of Perception Analyser}
\label{appendix:pa_output}
\end{figure*}

\begin{figure*}[t!]
\begin{tcolorbox}[title=Prompt provided to VLM to suggest grasp Keypoints]

You are an intelligent embodied agent that can perform interactive perception using visual and textual information. Your job is to analyse the provided image of a tabletop scene and suggest actions based on an input text question and task. The image is captured by a camera mounted on the wrist of Franka Panda robotic manipulator. 

Your goal is to suggest keypoints for the robotic arm to grasp the target object, with the ultimate aim of gaining more visual information to complete the input task or answer the input question.

\#\#\# Inputs Provided:

1. **Current Time Step**:

   \hspace{1cm}- **RGB Image**: A visual representation of the tabletop scene.
   
   \hspace{1cm}- **2D Grid Cube Image**: An image with a 5x5 grid overlay, annotated with x and y coordinates in the format [x; y]. This grid serves as the coordinate system for robotic arm actions.
   
   \hspace{1cm}- **Keypoints Image**: An RGB image with Keypoints overlay. The keypoints are annotated with letters "P" followed by a number
   
   \hspace{1cm}- **Optional RGB Image**: A visual representation of the bottom surface of the grasped object may be provided

2. **Historical Data**:

   - RGB images of the scene from all previous time steps, with the most recent towards the end.
   
   - Outputs from the previous time step:
   
     \hspace{1cm}- **Previous Action**: The action you suggested to perform the task.
     
     \hspace{1cm}- **Previous Scene Analysis**: Reasoning about the scene, including occlusions of the target object.
     
     \hspace{1cm}- **Previous Place Location**: The grid coordinate where the target object was placed.
     
     \hspace{1cm}- **Previous Contact Location**: The keypoint suggested for grasping the target object.
     
\#\#\# 2D Grid coordinate system:

    \hspace{1cm}- The grid's x and y coordinates represent positions within the 2D cube coordinate frame.
    
    \hspace{1cm}- Base of table: Lower limit of 2D cube's x-axis i.e (0,0 to 5,0)
    
    \hspace{1cm}- Top of table: Upper limit of 2D cube's x-axis i.e (0,5 to 5,5)
    
    \hspace{1cm}- Left of table: Upper limit of 2D cube's y-axis i.e (0,0 to 0,5)
    
    \hspace{1cm}- Right of table: Lower limit of 2D cube's y-axis i.e (5,0 to 5,5)
 
\#\#\# Task Objective:

Analyse the provided visual and textual data to:

    \hspace{1cm}1. Determine the best keypoint on the target object for the robotic arm to perform grasping. Choose only from keypoints **on the Target Object** you want to move.

    \hspace{1cm}2. Determine the best grid location for placing the target object.

    \hspace{1cm}3. Minimize disturbances to the scene while achieving the desired task outcome.

\#\#\# Key Instructions:

    \hspace{1cm}- output: You must suggest the best place location 
    
    \hspace{1cm}- Perform chain-of-thought reasoning
    
    \hspace{1cm}- Note that vertex\_x and vertex\_y are continuous float values rounded to one decimal place in the 2D cube coordinate frame
    
    \hspace{1cm}- Special Case for Time Step 0:
    
        \hspace{1.5cm}- If the current time step is `0`, it indicates the first iteration of the task.
        
        \hspace{1.5cm}- All previous outputs (Scene Analysis, Action, Place Location, and Contact Keypoint) will be empty.
        
    \hspace{1cm}- **Important**: The information from previous time steps indicates that the robotic task was complete, incomplete, or only partially successful. Use the provided textual and image data to generate an improved and complete solution for the task. Focus on precision, efficiency, and achieving the desired outcome.
        
\end{tcolorbox}
\caption{Prompt for grasping}
\label{appendix:prompt_grasp}
\end{figure*}

\begin{figure*}[t!]
\begin{tcolorbox}[title=Prompt provided to VLM for ZS-IP observe]
You are an intelligent embodied agent that can perfrom active perception using visual and textual information. Your job is to analyse the provided image of a tabletop scene based on an input task and text question. The image is captured by a camera mounted on the wrist of Franka Panda robotic manipulator. 

Your goal is to suggest camera positions and orientations for the robotic arm to find the target object, with the ultimate aim of gaining more visual information to complete the input task or answer the input question.

The input image will have a 3D cube with 3x3 grids overlaid on it. the grid's x and y vertices will be annotated in the format [x; y]. You must use the below conventions while proving the output

Base of table: Lower limit of 3D cube's x-axis i.e (0,0 to 5,0)

Top of table: Upper limit of 3D cube's x-axis i.e (0,5 to 5,5)

Layer Mapping: The 3D cube's z-axis consists of 3 discrete layers that are color coded. They represent the depth.

    \hspace{1cm}- vertex\_z = 0 (Blue layer) is at the table surface.
    
    \hspace{1cm}- vertex\_z = 1 (Black layer) is 10 cm above the table .
    
    \hspace{1cm}- vertex\_z = 2 (Green layer) is the topmost layer, 20 cm above the table.
    
Orientation Mapping: This is the fixed orientation of 35 degrees about x-axis (roll in our case).

    \hspace{1cm}- orient\_x = 0 when the target object of interest is flat on the table and perpendicular to the table 
    
    \hspace{1cm}- orient\_x = 1 for a +35° roll. Target is inclined towards the Base of Table
    
    \hspace{1cm}- orient\_x = 2 for a -35° roll. Target is inclined towards the Top of Table

You take actions by suggesting new camera positions and orientations with respect to 3D cube coordinate frame. The camera has an offset only along the z-axis and stops 35 cm above a given vertex\_z. You can suggest new meaningful camera positions and orientations to gain more visual information with the ultimate goal of being able to answer the input text question.

At each step you will be given your "State History" i.e position and orientation in the 3D cube's coordinate frame. You are encouraged to explore the environment while avoiding revisiting states by comparing with "State History".

Action Output consists of the next best camera location: 

    \hspace{1cm}- vertex\_x: Positive continuous value along x-axis in the cube coordinate frame rounded to one decimal
    
    \hspace{1cm}- vertex\_y: Positive continuous value along y-axis in the cube coordinate frame rounded to one decimal
    
    \hspace{1cm}- vertex\_z: Integer value along z-axis based on the Layer Mapping
    
    \hspace{1cm}- orient\_x: Integer value based on Orientation Mapping

\end{tcolorbox}
\caption{Prompt to perform camera in-hand movement}
\label{appendix:prmpt_cam_in_hand}
\end{figure*}

\end{document}